# Is Q-learning Provably Efficient?

Chi Jin*
University of California, Berkeley
chijin@cs.berkeley.edu

Zeyuan Allen-Zhu*
Microsoft Research, Redmond
zeyuan@csail.mit.edu

Sebastien Bubeck
Microsoft Research, Redmond
sebubeck@microsoft.com

Michael I. Jordan
University of California, Berkeley
jordan@cs.berkeley.edu

## Abstract

Model-free reinforcement learning (RL) algorithms, such as Q-learning, directly parameterize and update value functions or policies without explicitly modeling the environment. They are typically simpler, more flexible to use, and thus more prevalent in modern deep RL than model-based approaches. However, empirical work has suggested that model-free algorithms may require more samples to learn [7, 22]. The theoretical question of "whether model-free algorithms can be made *sample efficient*" is one of the most fundamental questions in RL, and remains unsolved even in the basic scenario with finitely many states and actions.

We prove that, in an episodic MDP setting, Q-learning with UCB exploration achieves regret $\tilde{\mathcal{O}}(\sqrt{H^3 SAT})$, where $S$ and $A$ are the numbers of states and actions, $H$ is the number of steps per episode, and $T$ is the total number of steps. This sample efficiency matches the optimal regret that can be achieved by any model-based approach, up to a single $\sqrt{H}$ factor. To the best of our knowledge, this is the first analysis in the model-free setting that establishes $\sqrt{T}$ regret *without* requiring access to a "simulator."

## 1 Introduction

Reinforcement Learning (RL) is a control-theoretic problem in which an agent tries to maximize its cumulative rewards via interacting with an unknown *environment* through time [26]. There are two main approaches to RL: model-based and model-free. Model-based algorithms make use of a model for the environment, forming a control policy based on this learned model. Model-free approaches dispense with the model and directly update the *value function*—the expected reward starting from each state, or the *policy*—the mapping from states to their subsequent actions. There has been a long debate on the relative pros and cons of the two approaches [7].

From the classical Q-learning algorithm [27] to modern DQN [17], A3C [18], TRPO [22], and others, most state-of-the-art RL has been in the model-free paradigm. Its pros—model-free algorithms are online, require less space, and, most importantly, are more expressive since specifying the value functions or policies is often more flexible than specifying the model for the environment—arguably outweigh its cons relative to model-based approaches. These relative advantages underly the significant successes of model-free algorithms in deep RL applications [17, 24].

On the other hand it is believed that model-free algorithms suffer from a higher sample complexity compared to model-based approaches. This has been evidenced empirically in [7, 22], and

---

*The first two authors contributed equally.

recent work has tried to improve the sample efficiency of model-free algorithms by combining them with model-based approaches [19, 21]. There is, however, little theory to support such blending, which requires a more quantitative understanding of relative sample complexities. Indeed, the following basic theoretical questions remain open:

**Can we design model-free algorithms that are sample efficient?**
In particular, **is Q-learning provably efficient?**

The answers remain elusive even in the basic tabular setting where the number of states and actions are finite. In this paper, we attack this problem head-on in the setting of the episodic Markov Decision Process (MDP) formalism (see Section 2 for a formal definition). In this setting, an episode consists of a run of MDP dynamics for $H$ steps, where the agent aims to maximize total reward over multiple episodes. We do not assume access to a "simulator" (which would allow us to query arbitrary state-action pairs of the MDP) and the agent is not allowed to "reset" within each episode. This makes our setting sufficiently challenging and realistic. In this setting, the standard Q-learning heuristic of incorporating $\varepsilon$-greedy exploration appears to take exponentially many episodes to learn [14].

As seen in the literature on bandits, the key to achieving good sample efficiency generally lies in managing the tradeoff between *exploration* and *exploitation*. One needs an efficient strategy to explore the uncertain environment while maximizing reward. In the model-based setting, a recent line of research has imported ideas from the bandit literature—including the use of upper confidence bounds (UCB) and improved design of exploration bonuses—and has obtained asymptotically optimal sample efficiency [1, 5, 10, 12]. In contrast, the understanding of model-free algorithms is still very limited. To the best of our knowledge, the only existing theoretical result on model-free RL that applies to the episodic setting is for *delayed Q-learning*; however, this algorithm is quite sample-inefficient compared to model-based approaches [25].

In this paper, we answer the two aforementioned questions affirmatively. We show that Q-learning, when equipped with a UCB exploration policy that incorporates estimates of the confidence of Q values and assign exploration bonuses, achieves total regret $\tilde{\mathcal{O}}(\sqrt{H^3SAT})$. Here, $S$ and $A$ are the numbers of states and actions, $H$ is the number of steps per episode, and $T$ is the total number of steps. Up to a $\sqrt{H}$ factor, our regret matches the information-theoretic optimum, which can be achieved by model-based algorithms [5, 12]. Since our algorithm is just Q-learning, it is online and does not store additional data besides the table of Q values (and a few integers per entry of this table). Thus, it also enjoys a significant advantage over model-based algorithms in terms of time and space complexities. To our best knowledge, this is the first sharp analysis for model-free algorithms—featuring $\sqrt{T}$ regret or equivalently $O(1/\varepsilon^2)$ samples for $\varepsilon$-optimal policy—*without* requiring access to a "simulator."

For practitioners, there are two key takeaways from our theoretical analysis:

1. The use of UCB exploration instead of $\varepsilon$-greedy exploration in the model-free setting allows for better treatment of uncertainties for different states and actions.

2. It is essential to use a learning rate which is $\alpha_t = O(H/t)$, instead of $1/t$, when a state-action pair is being updated for the $t$-th time. The former learning rate assigns more weight to updates that are more recent, as opposed to assigning uniform weights to all previous updates. This delicate choice of reweighting leads to the crucial difference between our sample-efficient guarantee versus earlier highly inefficient results that require exponentially many samples in $H$.

|  | **Algorithm** | **Regret** | **Time** | **Space** |
|---|---|---|---|---|
| Model-based | UCRL2 [10] [1] | at least $\tilde{\mathcal{O}}(\sqrt{H^4 S^2 AT})$ | $\Omega(TS^2A)$ | $\mathcal{O}(S^2AH)$ |
| | Agrawal and Jia [1] [1] | at least $\tilde{\mathcal{O}}(\sqrt{H^3 S^2 AT})$ | | |
| | UCBVI [5] [2] | $\tilde{\mathcal{O}}(\sqrt{H^2 SAT})$ | $\tilde{\mathcal{O}}(TS^2A)$ | |
| | vUCQ [12] [2] | $\tilde{\mathcal{O}}(\sqrt{H^2 SAT})$ | | |
| Model-free | Q-learning ($\varepsilon$-greedy) [14] (if 0 initialized) | $\Omega(\min\{T, A^{H/2}\})$ | $\mathcal{O}(T)$ | $\mathcal{O}(SAH)$ |
| | Delayed Q-learning [25] [3] | $\tilde{\mathcal{O}}_{S,A,H}(T^{4/5})$ | | |
| | Q-learning (UCB-H) | $\tilde{\mathcal{O}}(\sqrt{H^4 SAT})$ | | |
| | Q-learning (UCB-B) | $\tilde{\mathcal{O}}(\sqrt{H^3 SAT})$ | | |
| | lower bound | $\Omega(\sqrt{H^2 SAT})$ | - | - |

Table 1: Regret comparisons for RL algorithms on episodic MDP. $T = KH$ is totally number of steps, $H$ is the number of steps per episode, $S$ is the number of states, and $A$ is the number of actions. For clarity, this table is presented for $T \geq \text{poly}(S, A, H)$, omitting low order terms.

## 1.1 Related Work

In this section, we focus our attention on theoretical results for the tabular MDP setting, where the numbers of states and actions are finite. We acknowledge that there has been much recent work in RL for continuous state spaces [see, e.g., 9, 11], but this setting is beyond our scope.

**With simulator.** Some results assume access to a simulator [15] (a.k.a., a generative model [3]), which is a strong oracle that allows the algorithm to query arbitrary state-action pairs and return the reward and the next state. The majority of these results focus on an infinite-horizon MDP with discounted reward [e.g., 2, 3, 8, 16, 23]. When a simulator is available, model-free algorithms [2] (variants of Q-learning) are known to be almost as sample efficient as the best model-based algorithms [3]. However, the simulator setting is considered to much easier than standard RL, as it "does not require exploration" [2]. Indeed, a naive exploration strategy which queries all state-action pairs uniformly at random already leads to the most efficient algorithm for finding optimal policy [3].

**Without simulator.** Reinforcement learning becomes much more challenging without the presence of a simulator, and the choice of exploration policy can now determine the behavior of the learning algorithm. For instance, Q-learning with $\varepsilon$-greedy may take exponentially many episodes to learn the optimal policy [14] (for the sake of completeness, we present this result in our episodic language in Appendix A).

---

[1] Jaksch et al. [10] and Agrawal and Jia [1] apply to the more general setting of weakly communicating MDPs with $S'$ states and diameter $D$; our episodic MDP is a special case obtained by augmenting the state space so that $S' = SH$ and $D \geq H$.

[2] Azar et al. [5] and Kakade et al. [12] assume equal transition matrices $\mathbb{P}_1 = \cdots = \mathbb{P}_H$; in the setting of this paper $\mathbb{P}_1, \cdots, \mathbb{P}_H$ can be entirely different. This adds a factor of $\sqrt{H}$ to their total regret.

[3] Strehl et al. [25] applies to MDPs with $S'$ states and discount factor $\gamma$; our episodic MDP can be converted to that case by setting $S' = SH$ and $1 - \gamma = 1/H$. Their result only applies to the stochastic setting where initial states $x_1^k$ come from a fixed distribution, and only gives a PAC guarantee. We have translated it to a regret guarantee (see Section 3.1).

In the model-based setting, UCRL2 [10] and Agrawal and Jia [1] form estimates of the transition probabilities of the MDP using past samples, and add upper-confidence bounds (UCB) to the estimated transition matrix. When applying their results to the episodic MDP scenario, their total regret is at least $\tilde{\mathcal{O}}(\sqrt{H^4S^2AT})$ and $\tilde{\mathcal{O}}(\sqrt{H^3S^2AT})$ respectively.[1] In contrast, the information-theoretic lower bound is $\tilde{\mathcal{O}}(\sqrt{H^2SAT})$. The additional $\sqrt{S}$ and $\sqrt{H}$ factors were later removed by the UCBVI algorithm [5] which adds a UCB bonus directly to the Q values instead of the estimated transition matrix.[2] The vUCQ algorithm [12] is similar to UCBVI but improves lower-order regret terms using variance reduction.

We note that despite the sharp regret guarantees, all of the results in this line of research require estimating and storing the entire transition matrix and thus suffer from unfavorable time and space complexities compared to model-free algorithms.

In the model-free setting, Strehl et al. [25] introduced delayed Q-learning, where, to find an $\varepsilon$-optimal policy, the Q value for each state-action pair is updated only once every $m = \tilde{\mathcal{O}}(1/\varepsilon^2)$ times this pair is visited. In contrast to the incremental update of Q-learning, delayed Q-learning always replaces old Q values with the average of the most recent $m$ experiences. When translated to the setting of this paper, this gives $\tilde{\mathcal{O}}(T^{4/5})$ total regret, ignoring factors in $S, A$ and $H$.[3] This is quite suboptimal compared to the $\tilde{\mathcal{O}}(\sqrt{T})$ regret achieved by model-based algorithm.

## 2   Preliminary

We consider the setting of a tabular episodic Markov decision process, $\mathrm{MDP}(\mathcal{S}, \mathcal{A}, \mathrm{H}, \mathbb{P}, \mathrm{r})$, where $\mathcal{S}$ is the set of states with $|\mathcal{S}| = S$, $\mathcal{A}$ is the set of actions with $|\mathcal{A}| = A$, $H$ is the number of steps in each episode, $\mathbb{P}$ is the transition matrix so that $\mathbb{P}_h(\cdot|x, a)$ gives the distribution over states if action $a$ is taken for state $x$ at step $h \in [H]$, and $r_h \colon \mathcal{S} \times \mathcal{A} \to [0, 1]$ is the deterministic reward function at step $h$.[4]

In each episode of this MDP, an initial state $x_1$ is picked arbitrarily by an adversary. Then, at each step $h \in [H]$, the agent observes state $x_h \in \mathcal{S}$, picks an action $a_h \in \mathcal{A}$, receives reward $r_h(x_h, a_h)$, and then transitions to a next state, $x_{h+1}$, that is drawn from the distribution $\mathbb{P}_h(\cdot|x_h, a_h)$. The episode ends when $x_{H+1}$ is reached.

A policy $\pi$ of an agent is a collection of $H$ functions $\{\pi_h \colon \mathcal{S} \to \mathcal{A}\}_{h \in [H]}$. We use $V_h^\pi \colon \mathcal{S} \to \mathbb{R}$ to denote the value function at step $h$ under policy $\pi$, so that $V_h^\pi(x)$ gives the expected sum of remaining rewards received under policy $\pi$, starting from $x_h = x$, until the end of the episode. In symbols:

$$V_h^\pi(x) := \mathbb{E}\left[\sum_{h'=h}^H r_{h'}(x_{h'}, \pi_{h'}(x_{h'}))|x_h = x\right] \ .$$

Accordingly, we also define $Q_h^\pi \colon \mathcal{S} \times \mathcal{A} \to \mathbb{R}$ to denote $Q$-value function at step $h$ so that $Q_h^\pi(x, a)$ gives the expected sum of remaining rewards received under policy $\pi$, starting from $x_h = x, a_h = a$, till the end of the episode. In symbols:

$$Q_h^\pi(x, a) := r_h(x, a) + \mathbb{E}[\sum_{h'=h+1}^H r_{h'}(x_{h'}, \pi_{h'}(x_{h'}))|x_h = x, a_h = a] \ .$$

Since the state and action spaces, and the horizon, are all finite, there always exists (see, e.g., [5]) an optimal policy $\pi^\star$ which gives the optimal value $V_h^\star(x) = \sup_\pi V_h^\pi(x)$ for all $x \in \mathcal{S}$ and $h \in [H]$. For simplicity, we denote $[\mathbb{P}_h V_{h+1}](x, a) := \mathbb{E}_{x' \sim \mathbb{P}(\cdot|x,a)} V_{h+1}(x')$. Recall the Bellman equation and

---

[4]While we study deterministic reward functions for notational simplicity, our results generalize to randomized reward functions. Also, we assume the reward is in $[0, 1]$ without loss of generality.

**Algorithm 1** Q-learning with UCB-Hoeffding

1: initialize $Q_h(x, a) \leftarrow H$ and $N_h(x, a) \leftarrow 0$ for all $(x, a, h) \in \mathcal{S} \times \mathcal{A} \times [H]$.
2: **for** episode $k = 1, \ldots, K$ **do**
3:     receive $x_1$.
4:     **for** step $h = 1, \ldots, H$ **do**
5:         Take action $a_h \leftarrow \mathrm{argmax}_{a'} Q_h(x_h, a')$, and observe $x_{h+1}$.
6:         $t = N_h(x_h, a_h) \leftarrow N_h(x_h, a_h) + 1;\ \ b_t \leftarrow c\sqrt{H^3\iota/t}$.
7:         $Q_h(x_h, a_h) \leftarrow (1 - \alpha_t)Q_h(x_h, a_h) + \alpha_t[r_h(x_h, a_h) + V_{h+1}(x_{h+1}) + b_t]$.
8:         $V_h(x_h) \leftarrow \min\{H, \max_{a' \in \mathcal{A}} Q_h(x_h, a')\}$.

the Bellman optimality equation:

$$\begin{cases} V_h^\pi(x) = Q_h^\pi(x, \pi_h(x)) \\ Q_h^\pi(x, a) := (r_h + \mathbb{P}_h V_{h+1}^\pi)(x, a) \\ V_{H+1}^\pi(x) = 0 \quad \forall x \in \mathcal{S} \end{cases} \text{ and } \begin{cases} V_h^\star(x) = \max_{a \in \mathcal{A}} Q_h^\star(x, a) \\ Q_h^\star(x, a) := (r_h + \mathbb{P}_h V_{h+1}^\star)(x, a) \\ V_{H+1}^\star(x) = 0 \quad \forall x \in \mathcal{S} \ . \end{cases} \quad (2.1)$$

The agent plays the game for $K$ episodes $k = 1, 2, \ldots, K$, and we let the adversary pick a starting state $x_1^k$ for each episode $k$, and let the agent choose a policy $\pi_k$ before starting the $k$-th episode. The total (expected) regret is then

$$\mathrm{Regret}(K) = \sum_{k=1}^K \left[ V_1^\star(x_1^k) - V_1^{\pi_k}(x_1^k) \right] \ .$$

## 3 Main Results

In this section, we present our main theoretical result—a sample complexity result for a variant of Q-learning that incorporates UCB exploration. We also present a theorem that establishes an information-theoretic lower bound for episodic MDP.

As seen in the bandit setting, the choice of exploration policy plays an essential role in the efficiency of a learning algorithm. In episodic MDP, Q-learning with the commonly used $\varepsilon$-greedy exploration strategy can be very inefficient: it can take exponentially many episodes to learn [14] (see also Appendix A). In contrast, our algorithm (Algorithm 1), which is Q-learning with an upper-confidence bound (UCB) exploration strategy, will be seen to be efficient. This algorithm maintains Q values, $Q_h(x, a)$, for all $(x, a, h) \in \mathcal{S} \times \mathcal{A} \times [H]$ and the corresponding V values $V_h(x) \leftarrow \min\{H, \max_{a' \in \mathcal{A}} Q_h(x, a')\}$. If, at time step $h \in [H]$, the state is $x \in \mathcal{S}$, the algorithm takes the action $a \in \mathcal{A}$ that maximizes the current estimate $Q_h(x, a)$, and is apprised of the next state $x' \in \mathcal{S}$. The algorithm then updates the Q values:

$$Q_h(x, a) \leftarrow (1 - \alpha_t)Q_h(x, a) + \alpha_t[r_h(x, a) + V_{h+1}(x') + b_t] \ ,$$

where $t$ is the counter for how many times the algorithm has visited the state-action pair $(x, a)$ at step $h$, $b_t$ is the confidence bonus indicating how certain the algorithm is about current state-action pair, and $\alpha_t$ is a learning rate defined as follows:

$$\alpha_t := \frac{H + 1}{H + t} \ . \quad (3.1)$$

As mentioned in the introduction, our choice of learning rate $\alpha_t$ scales as $O(H/t)$ instead of $O(1/t)$—this is crucial to obtain regret that is not exponential in $H$.

We present analyses for two different specifications of the upper confidence bonus $b_t$ in this

paper:

**Q-learning with Hoeffding-style bonus.** The first (and simpler) choice is $b_t = O(\sqrt{H^3\iota/t})$. (Here, and throughout this paper, we use $\iota := \log(SAT/p)$ to denote a log factor.) This choice of bonus makes sense intuitively because: (1) Q-values are upper-bounded by $H$, and, accordingly, (2) Hoeffding-type martingale concentration inequalities imply that if we have visited $(x, a)$ for $t$ times, then a confidence bound for the Q value scales as $1/\sqrt{t}$. For this reason, we call this choice *UCB-Hoeffding* (UCB-H). See Algorithm 1.

**Theorem 1** (Hoeffding). *There exists an absolute constant $c > 0$ such that, for any $p \in (0,1)$, if we choose $b_t = c\sqrt{H^3\iota/t}$, then with probability $1-p$, the total regret of Q-learning with UCB-Hoeffding (see Algorithm 1) is at most $O(\sqrt{H^4SAT\iota})$, where $\iota := \log(SAT/p)$.*

Theorem 1 shows, under a rather simple choice of exploration bonus, Q-learning can be made very efficient, enjoying a $\tilde{\mathcal{O}}(\sqrt{T})$ regret which is optimal in terms of dependence on $T$. To the best of our knowledge, this is the first analysis of a model-free procedure that features a $\sqrt{T}$ regret *without* requiring access to a "simulator."

Compared to the previous model-based results, Theorem 1 shows that the regret (or equivalently the sample complexity; see discussion in Section 3.1) of this version of Q-learning is as good as the best model-based one in terms of the dependency on the number of states $S$, actions $A$ and the total number of steps $T$. Although our regret slightly increases the dependency on $H$, the algorithm is online and does not store additional data besides the table of Q values (and a few integers per entry of this table). Thus, it enjoys an advantage over model-based algorithms in time and space complexities, especially when the number of states $S$ is large.

**Q-learning with Bernstein-style bonus.** Our second specification of $b_t$ makes use of a Bernstein-style upper confidence bound. The key observation is that, although in the worst case the value function is at most $H$ for any state-action pair, if we sum up the "total variance of the value function" for an entire episode, we obtain a factor of only $O(H^2)$ as opposed to the naive $O(H^3)$ bound (see Lemma C.5). This implies that the use of a Bernstein-type martingale concentration result could be sharper than the Hoeffding-type bound by an additional factor of $H$.[5] (The idea of using Bernstein instead of Hoeffding for reinforcement learning applications has appeared in previous work; see, e.g., [3, 4, 16].)

Using Bernstein concentration requires us to design the bonus term $b_t$ more carefully, as it now depends on the empirical variance of $V_{h+1}(x')$ where $x'$ is the next state over the previous $t$ visits of current state-action $(x, a)$. This empirical variance can be computed in an online fashion without increasing the space complexity of Q-learning. We defer the full specification of $b_t$ to Algorithm 2 in Appendix C. We now state the regret theorem for this approach.

**Theorem 2** (Bernstein). *For any $p \in (0,1)$, one can specify $b_t$ so that with probability $1-p$, the total regret of Q-learning with UCB-Bernstein (see Algorithm 2) is at most $O(\sqrt{H^3SAT\iota} + \sqrt{H^9S^3A^3}\cdot\iota^2)$.*

Theorem 2 shows that for Q-learning with UCB-B exploration, the leading term in regret (which scales as $\sqrt{T}$) improves by a factor of $\sqrt{H}$ over UCB-H exploration, at the price of using a more complicated exploration bonus design. The asymptotic regret of UCB-B is now only one $\sqrt{H}$ factor worse than the best regret achieved by model-based algorithms.

---
[5]Recall that for independent zero-mean random variables $X_1, \ldots, X_T$ satisfying $|X_i| \leq M$, their summation does not exceed $\tilde{\mathcal{O}}(M\sqrt{T})$ with high probability using Hoeffding concentration. If we have in hand a better variance bound, this can be improved to $\tilde{\mathcal{O}}(M + \sqrt{\sum_i \mathbb{E}[X_i]^2})$ using Bernstein concentration.

We also note that Theorem 2 has an additive term $O(\sqrt{H^9 S^3 A^3} \cdot \iota^2)$ in its regret, which dominates the total regret when $T$ is not very large compared with $S, A$ and $H$. It is not clear whether this lower-order term is essential, or is due to technical aspects of the current analysis.

**Information-theoretical limit.** To demonstrate the sharpness of our results, we also note an information-theoretic lower bound for the episodic MDP setting studied in this paper:

**Theorem 3.** *For the episodic MDP problem studied in this paper, the expected regret for any algorithm must be at least $\Omega(\sqrt{H^2 SAT})$.*

Theorem 3 (see Appendix D for details) shows that both variants of our algorithm are nearly optimal, in the sense they differ from the optimal regret by a factor of $H$ and $\sqrt{H}$, respectively.

## 3.1 From Regret to PAC Guarantee

Recall that the probably approximately correct (PAC) learning setting for RL provides sample complexity guarantee to find a near-optimal policy [13]. In this setting, the initial state $x_1 \in \mathcal{S}$ is sampled from a fixed initial distribution, rather than being chosen adversarially. Without loss of generality, we only discuss here the case in which $x_1$ is fixed; the general case reduces to this case by adding an additional time step at the beginning of each episode. The PAC-learning question is "how many samples are needed to find an $\varepsilon$-optimal policy $\pi$ satisfying $V_1^\star(x_1) - V_1^\pi(x_1) \leq \varepsilon$?"

Any algorithm with total regret sublinear in $T$ yields a finite sample complexity in the PAC setting. Indeed, suppose we have total regret $\sum_{k=1}^{K} [V_1^\star(x_1) - V_1^{\pi_k}(x_1)] \leq C \cdot T^{1-\alpha}$, where $\alpha \in (0,1)$ is a absolute constant, and $C$ is independent of $T$. Then, by randomly selecting $\pi = \pi_k$ for $k = 1, 2, \ldots, K$, we have $V_1^\star(x_1) - V_1^\pi(x_1) \leq 3CH \cdot T^{-\alpha}$ with probability at least $2/3$. Therefore, for every $\varepsilon \in (0, H]$, our Theorem 1 (for UCB-H) and Theorem 2 (for UCB-B) also find $\varepsilon$-optimal policies in the PAC setting using $\tilde{\mathcal{O}}(H^5 SA/\varepsilon^2)$ and $\tilde{\mathcal{O}}(H^4 SA/\varepsilon^2)$ samples respectively.

Conversely, any algorithm with finite sample complexity in the PAC setting translates to sublinear total regret in non-adversarial case (assuming $x_1$ is chosen from a fixed distribution). Suppose the algorithm finds $\varepsilon$-optimal policy $\pi$ using $T_1 = C \cdot \varepsilon^{-\beta}$ samples where $\beta \geq 1$ is a constant. Then, we can use this $\pi$ to play the game for another $T - T_1$ steps, giving total regret $T_1 + \varepsilon(T - T_1)/H$. After balancing $T$ and $T_1$ optimally, this gives $\tilde{\mathcal{O}}\big(C^{1+\beta} \cdot (T/H)^{\beta/(1+\beta)}\big)$ total regret. For instance, Strehl et al. [25] gives sampling complexity $\propto 1/\varepsilon^4$ in the PAC setting, and this translates to $\propto T^{4/5}$ total regret.

# 4 Proof for Q-learning with UCB-Hoeffding

In this section, we provide the full proof of Theorem 1. Intuitively, the episodic MDP with $H$ steps per epsiode can be viewed as a contextual bandit of $H$ "layers." The key challenge here is to control the way error and confidence propagate through different "layers" in an online fashion, where our specific choice of exploration bonus and learning rate make the regret as sharp as possible.

**Notation.** We denote by $\mathbb{I}[A]$ the indicator function for event $A$. We denote by $(x_h^k, a_h^k)$ the actual state-action pair observed and chosen at step $h$ of episode $k$. We also denote by $Q_h^k, V_h^k, N_h^k$ respectively the $Q_h, V_h, N_h$ functions at the *beginning* of episode $k$. Using this notation, the update equation at episode $k$ can be rewritten as follows, for every $h \in [H]$:

$$Q_h^{k+1}(x, a) = \begin{cases} (1 - \alpha_t) Q_h^k(x, a) + \alpha_t [r_h(x, a) + V_{h+1}^k(x_{h+1}^k) + b_t] & \text{if } (x, a) = (x_h^k, a_h^k) \\ Q_h^k(x, a) & \text{otherwise} \end{cases} \quad (4.1)$$

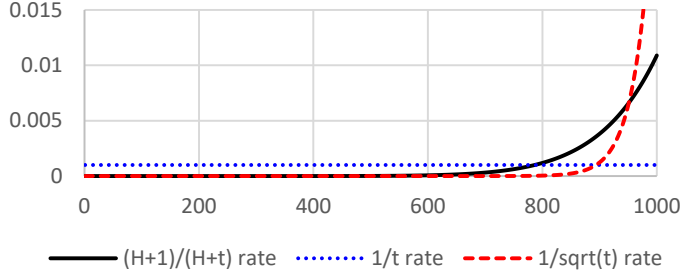

Figure 1: Illustration of $\{\alpha^i_{1000}\}_{i=1}^{1000}$ for learning rates $\alpha_t = \frac{H+1}{H+t}, \frac{1}{t}$ and $\frac{1}{\sqrt{t}}$ when $H = 10$.

Accordingly,
$$V_h^k(x) \leftarrow \min\left\{H, \max_{a' \in \mathcal{A}} Q_h^k(x, a')\right\}, \quad \forall x \in \mathcal{S} \ .$$

Recall that we have $[\mathbb{P}_h V_{h+1}](x, a) := \mathbb{E}_{x' \sim \mathbb{P}_h(\cdot|x,a)} V_{h+1}(x')$. We also denote its empirical counterpart of episode $k$ as $[\hat{\mathbb{P}}_h^k V_{h+1}](x, a) := V_{h+1}(x_{h+1}^k)$, which is defined only for $(x, a) = (x_h^k, a_h^k)$.

Recall that we have chosen the learning rate as $\alpha_t := \frac{H+1}{H+t}$. For notational convenience, we also introduce the following related quantities:
$$\alpha_t^0 = \prod_{j=1}^t (1 - \alpha_j), \qquad \alpha_t^i = \alpha_i \prod_{j=i+1}^t (1 - \alpha_j) \ . \tag{4.2}$$

It is easy to verify that (1) $\sum_{i=1}^t \alpha_t^i = 1$ and $\alpha_t^0 = 0$ for $t \geq 1$; (2) $\sum_{i=1}^t \alpha_t^i = 0$ and $\alpha_t^0 = 1$ for $t = 0$.

**Favoring Later Updates.** At any $(x, a, h, k) \in \mathcal{S} \times \mathcal{A} \times [H] \times [K]$, let $t = N_h^k(x, a)$ and suppose $(x, a)$ was previously taken at step $h$ of episodes $k_1, \ldots, k_t < k$. By the update equation (4.1) and the definition of $\alpha_t^i$ in (4.2), we have:
$$Q_h^k(x, a) = \alpha_t^0 H + \sum_{i=1}^t \alpha_t^i \left[ r_h(x, a) + V_{h+1}^{k_i}(x_{h+1}^{k_i}) + b_i \right] \ . \tag{4.3}$$

According to (4.3), the $Q$ value at episode $k$ equals a weighted average of the $V$ values of the "next states" with weights $\alpha_t^1, \ldots, \alpha_t^t$. As one can see from Figure 1, our choice of the learning rate $\alpha_t = \frac{H+1}{H+t}$ ensures that, approximately speaking, the last $1/H$ fraction of the indices $i$ is given non-negligible weights, whereas the first $1 - 1/H$ fraction is forgotten. This ensures that the information accumulates smoothly across the $H$ layers of the MDP. If one were to use $\alpha_t = \frac{1}{t}$ instead, the weights $\alpha_t^1, \ldots, \alpha_t^t$ would all equal $1/t$, and using those $V$ values from earlier episodes would hurt the accuracy of the $Q$ function. In contrast, if one were to use $\alpha_t = 1/\sqrt{t}$ instead, the weights $\alpha_t^1, \ldots, \alpha_t^t$ would concentrate too much on the most recent episodes, which would incur high variance.

### 4.1 Proof Details

We first present an auxiliary lemma which exhibits some important properties that result from our choice of learning rate. The proof is based on simple manipulations on the definition of $\alpha_t$, and is provided in Appendix B.

**Lemma 4.1.** *The following properties hold for $\alpha_t^i$:*

(a) $\frac{1}{\sqrt{t}} \leq \sum_{i=1}^t \frac{\alpha_t^i}{\sqrt{i}} \leq \frac{2}{\sqrt{t}}$ *for every $t \geq 1$.*

(b) $\max_{i \in [t]} \alpha_t^i \le \frac{2H}{t}$ and $\sum_{i=1}^{t}(\alpha_t^i)^2 \le \frac{2H}{t}$ for every $t \ge 1$.

(c) $\sum_{t=i}^{\infty} \alpha_t^i = 1 + \frac{1}{H}$ for every $i \ge 1$.

We note that property (c) is especially important—as we will show later, each step in one episode can blow up the regret by a multiplicative factor of $\sum_{t=i}^{\infty} \alpha_t^i$. With our choice of learning rate, we ensure that this blow-up is at most $(1 + 1/H)^H$, which is a constant factor.

We nnow proceed to the formal proof. We start with a lemma that gives a recursive formula for $Q - Q^\star$, as a weighted average of previous updates.

**Lemma 4.2** (recursion on $Q$). *For any $(x, a, h) \in \mathcal{S} \times \mathcal{A} \times [H]$ and episode $k \in [K]$, let $t = N_h^k(x, a)$ and suppose $(x, a)$ was previously taken at step $h$ of episodes $k_1, \ldots, k_t < k$. Then:*

$$(Q_h^k - Q_h^\star)(x, a) = \alpha_t^0 (H - Q_h^\star(x, a)) + \sum_{i=1}^{t} \alpha_t^i \left[ (V_{h+1}^{k_i} - V_{h+1}^\star)(x_{h+1}^{k_i}) + [(\hat{\mathbb{P}}_h^{k_i} - \mathbb{P}_h) V_{h+1}^\star](x, a) + b_i \right] .$$

*Proof of Lemma 4.2.* From the Bellman optimality equation, $Q_h^\star(x, a) = (r_h + \mathbb{P}_h V_{h+1}^\star)(x, a)$, our notation $[\hat{\mathbb{P}}_h^{k_i} V_{h+1}](x, a) := V_{h+1}(x_{h+1}^{k_i})$, and the fact that $\sum_{i=0}^{t} \alpha_t^i = 1$, we have

$$Q_h^\star(x, a) = \alpha_t^0 Q_h^\star(x, a) + \sum_{i=1}^{t} \alpha_t^i \left[ r_h(x, a) + (\mathbb{P}_h - \hat{\mathbb{P}}_h^{k_i}) V_{h+1}^\star(x, a) + V_{h+1}^\star(x_{h+1}^{k_i}) \right] .$$

Subtracting the formula (4.3) from this equation, we obtain Lemma 4.2. $\square$

Next, using Lemma 4.2 and the Azuma-Hoeffding concentration bound, our next lemma shows that $Q^k$ is always an upper bound on $Q^\star$ at any episode $k$, and the difference between $Q^k$ and $Q^\star$ can be bounded by quantities from the next step.

**Lemma 4.3** (bound on $Q^k - Q^\star$). *There exists an absolute constant $c > 0$ such that, for any $p \in (0, 1)$, letting $b_t = c\sqrt{H^3 \iota / t}$, we have $\beta_t = 2 \sum_{i=1}^{t} \alpha_t^i b_i \le 4c\sqrt{H^3 \iota / t}$ and, with probability at least $1 - p$, the following holds simultaneously for all $(x, a, h, k) \in \mathcal{S} \times \mathcal{A} \times [H] \times [K]$:*

$$0 \le (Q_h^k - Q_h^\star)(x, a) \le \alpha_t^0 H + \sum_{i=1}^{t} \alpha_t^i (V_{h+1}^{k_i} - V_{h+1}^\star)(x_{h+1}^{k_i}) + \beta_t ,$$

*where $t = N_h^k(x, a)$ and $k_1, \ldots, k_t < k$ are the episodes where $(x, a)$ was taken at step $h$.*

*Proof of Lemma 4.3.* For each fixed $(x, a, h) \in \mathcal{S} \times \mathcal{A} \times [H]$, let us denote $k_0 = 0$, and denote

$$k_i = \min \left( \{ k \in [K] \mid k > k_{i-1} \wedge (x_h^k, a_h^k) = (x, a) \} \cup \{K + 1\} \right) .$$

That is, $k_i$ is the episode of which $(x, a)$ was taken at step $h$ for the $i$th time (or $k_i = K + 1$ if it is taken for fewer than $i$ times). The random variable $k_i$ is clearly a stopping time. Let $\mathcal{F}_i$ be the $\sigma$-field generated by all the random variables until episode $k_i$, step $h$. Then, $\left( \mathbb{I}[k_i \le K] \cdot [(\hat{\mathbb{P}}_h^{k_i} - \mathbb{P}_h) V_{h+1}^\star](x, a) \right)_{i=1}^{\tau}$ is a martingale difference sequence w.r.t the filtration $\{\mathcal{F}_i\}_{i \ge 0}$. By Azuma-Hoeffding and a union bound, we have that with probability at least $1 - p/(SAH)$:

$$\forall \tau \in [K] : \quad \left| \sum_{i=1}^{\tau} \alpha_\tau^i \cdot \mathbb{I}[k_i \le K] \cdot [(\hat{\mathbb{P}}_h^{k_i} - \mathbb{P}_h) V_{h+1}^\star](x, a) \right| \le \frac{cH}{2} \sqrt{\sum_{i=1}^{\tau} (\alpha_\tau^i)^2 \cdot \iota} \le c\sqrt{\frac{H^3 \iota}{\tau}} , \quad (4.4)$$

for some absolute constant $c$. Because inequality (4.4) holds for all fixed $\tau \in [K]$ uniformly, it also holds for $\tau = t = N_h^k(x, a) \le K$, which is a random variable, where $k \in [K]$. Also note $\mathbb{I}[k_i \le K] = 1$ for all $i \leq N_h^k(x, a)$. Putting everything together, and using a union bound, we see that with least $1 - p$ probability, the following holds simultaneously for all $(x, a, h, k) \in \mathcal{S} \times \mathcal{A} \times [H] \times [K]$:

$$\left| \sum_{i=1}^{t} \alpha_t^i [(\hat{\mathbb{P}}_h^{k_i} - \mathbb{P}_h) V_{h+1}^\star](x, a) \right| \leq c \sqrt{\frac{H^3 \iota}{t}} \quad \text{where} \quad t = N_h^k(x, a) \ . \tag{4.5}$$

On the other hand, if we choose $b_t = c\sqrt{H^3 \iota/t}$ for the same constant $c$ in Eq. (4.4), we have $\beta_t/2 = \sum_{i=1}^{t} \alpha_t^i b_i \in [c\sqrt{H^3 \iota/t}, 2c\sqrt{H^3 \iota/t}]$ according to Lemma 4.1.a. Then the right-hand side of Lemma 4.3 follows immediately from Lemma 4.2 and inequality (4.5). The left-hand side also follows from Lemma 4.2 and Eq. (4.5) and induction on $h = H, H - 1, \ldots, 1$. □

We are now ready to prove Theorem 1. The proof decomposes the regret in a recursive form, and carefully controls the error propagation with repeated usage of Lemma 4.3.

*Proof of Theorem 1.* Denote by

$$\delta_h^k := (V_h^k - V_h^{\pi_k})(x_h^k) \quad \text{and} \quad \phi_h^k := (V_h^k - V_h^\star)(x_h^k) \ .$$

By Lemma 4.3, we have that with $1 - p$ probability, $Q_h^k \geq Q_h^\star$ and thus $V_h^k \geq V_h^\star$. Thus, the total regret can be upper bounded:

$$\text{Regret}(K) = \sum_{k=1}^{K} (V_1^\star - V_1^{\pi_k})(x_1^k) \leq \sum_{k=1}^{K} (V_1^k - V_1^{\pi_k})(x_1^k) = \sum_{k=1}^{K} \delta_1^k \ .$$

The main idea of the rest of the proof is to upper bound $\sum_{k=1}^{K} \delta_h^k$ by the next step $\sum_{k=1}^{K} \delta_{h+1}^k$, thus giving a recursive formula to calculate total regret. We can obtain such a recursive formula by relating $\sum_{k=1}^{K} \delta_h^k$ to $\sum_{k=1}^{K} \phi_h^k$.

For any fixed $(k, h) \in [K] \times [H]$, let $t = N_h^k(x_h^k, a_h^k)$, and suppose $(x_h^k, a_h^k)$ was previously taken at step $h$ of episodes $k_1, \ldots, k_t < k$. Then we have:

$$\delta_h^k = (V_h^k - V_h^{\pi_k})(x_h^k) \overset{\text{①}}{\leq} (Q_h^k - Q_h^{\pi_k})(x_h^k, a_h^k)$$
$$= (Q_h^k - Q_h^\star)(x_h^k, a_h^k) + (Q_h^\star - Q_h^{\pi_k})(x_h^k, a_h^k)$$
$$\overset{\text{②}}{\leq} \alpha_t^0 H + \sum_{i=1}^{t} \alpha_t^i \phi_{h+1}^{k_i} + \beta_t + [\mathbb{P}_h (V_{h+1}^\star - V_{h+1}^{\pi_k})](x_h^k, a_h^k)$$
$$\overset{\text{③}}{=} \alpha_t^0 H + \sum_{i=1}^{t} \alpha_t^i \phi_{h+1}^{k_i} + \beta_t - \phi_{h+1}^k + \delta_{h+1}^k + \xi_{h+1}^k \ , \tag{4.6}$$

where $\beta_t = 2 \sum \alpha_t^i b_i \leq O(1) \sqrt{H^3 \iota/t}$ and $\xi_{h+1}^k := [(\mathbb{P}_h - \hat{\mathbb{P}}_h^k)(V_{h+1}^\star - V_{h+1}^k)](x_h^k, a_h^k)$ is a martingale difference sequence. Inequality ① holds because $V_h^k(x_h^k) \leq \max_{a' \in \mathcal{A}} Q_h^k(x_h^k, a') = Q_h^k(x_h^k, a_h^k)$, and inequality ② holds by Lemma 4.3 and the Bellman equation (2.1). Finally, equality ③ holds by definition $\delta_{h+1}^k - \phi_{h+1}^k = (V_{h+1}^\star - V_{h+1}^{\pi_k})(x_{h+1}^k)$.

We turn to computing the summation $\sum_{k=1}^{K} \delta_h^k$. Denoting by $n_h^k = N_h^k(x_h^k, a_h^k)$, we have:

$$\sum_{k=1}^{K} \alpha_{n_h^k}^0 H = \sum_{k=1}^{K} H \cdot \mathbb{I}[n_h^k = 0] \leq SAH \ .$$

The key step is to upper bound the second term in (4.6), which is:

$$\sum_{k=1}^{K} \sum_{i=1}^{n_h^k} \alpha_{n_h^k}^i \phi_{h+1}^{k_i(x_h^k, a_h^k)},$$

where $k_i(x_h^k, a_h^k)$ is the episode in which $(x_h^k, a_h^k)$ was taken at step $h$ for the $i$th time. We regroup the summands in a different way. For every $k' \in [K]$, the term $\phi_{h+1}^{k'}$ appears in the summand with $k > k'$ if and only if $(x_h^k, s_h^k) = (x_h^{k'}, s_h^{k'})$. The first time it appears we have $n_h^k = n_h^{k'} + 1$, the second time it appears we have $n_h^k = n_h^{k'} + 2$, and so on. Therefore

$$\sum_{k=1}^{K} \sum_{i=1}^{n_h^k} \alpha_{n_h^k}^i \phi_{h+1}^{k_i(x_h^k, a_h^k)} \leq \sum_{k'=1}^{K} \phi_{h+1}^{k'} \sum_{t=n_h^{k'}+1}^{\infty} \alpha_t^{n_h^{k'}} \leq \left(1 + \frac{1}{H}\right) \sum_{k=1}^{K} \phi_{h+1}^k,$$

where the final inequality uses $\sum_{t=i}^{\infty} \alpha_t^i = 1 + \frac{1}{H}$ from Lemma 4.1.c. Plugging these back into (4.6), we have:

$$\sum_{k=1}^{K} \delta_h^k \leq SAH + \left(1 + \frac{1}{H}\right) \sum_{k=1}^{K} \phi_{h+1}^k - \sum_{k=1}^{K} \phi_{h+1}^k + \sum_{k=1}^{K} \delta_{h+1}^k + \sum_{k=1}^{K} (\beta_{n_h^k} + \xi_{h+1}^k)$$

$$\leq SAH + \left(1 + \frac{1}{H}\right) \sum_{k=1}^{K} \delta_{h+1}^k + \sum_{k=1}^{K} (\beta_{n_h^k} + \xi_{h+1}^k) , \qquad (4.7)$$

where the final inequality uses $\phi_{h+1}^k \leq \delta_{h+1}^k$ (owing to the fact that $V^\star \geq V^{\pi_k}$). Recursing the result for $h = 1, 2, \ldots, H$, and using the fact $\delta_{H+1}^K \equiv 0$, we have:

$$\sum_{k=1}^{K} \delta_1^k \leq O\Big(H^2 SA + \sum_{h=1}^{H} \sum_{k=1}^{K} (\beta_{n_h^k} + \xi_{h+1}^k)\Big).$$

Finally, by the pigeonhole principle, for any $h \in [H]$:

$$\sum_{k=1}^{K} \beta_{n_h^k} \leq O(1) \cdot \sum_{k=1}^{K} \sqrt{\frac{H^3 \iota}{n_h^k}} = O(1) \cdot \sum_{x,a} \sum_{n=1}^{N_h^K(x,a)} \sqrt{\frac{H^3 \iota}{n}} \overset{①}{\leq} O\big(\sqrt{H^3 SAK\iota}\big) = O\big(\sqrt{H^2 SAT\iota}\big) \quad (4.8)$$

where inequality ① is true because $\sum_{x,a} N_h^K(x,a) = K$ and the left-hand side of ① is maximized when $N_h^K(x,a) = K/SA$ for all $x, a$. Also, by the Azuma-Hoeffding inequality, with probability $1 - p$, we have:

$$\Big|\sum_{h=1}^{H} \sum_{k=1}^{K} \xi_{h+1}^k\Big| = \Big|\sum_{h=1}^{H} \sum_{k=1}^{K} [(\mathbb{P}_h - \hat{\mathbb{P}}_h^k)(V_{h+1}^\star - V_{h+1}^k)](x_h^k, a_h^k)\Big| \leq cH\sqrt{T\iota}.$$

This establishes $\sum_{k=1}^{K} \delta_1^k \leq O\big(H^2 SA + \sqrt{H^4 SAT\iota}\big)$. We note that when $T \geq \sqrt{H^4 SAT\iota}$, we have $\sqrt{H^4 SAT\iota} \geq H^2 SA$, and when $T \leq \sqrt{H^4 SAT\iota}$, we have $\sum_{k=1}^{K} \delta_1^k \leq HK = T \leq \sqrt{H^4 SAT\iota}$. Therefore, we can remove the $H^2 SA$ term in the regret upper bound.

In sum, we have $\sum_{k=1}^{K} \delta_1^k \leq O\big(H^2 SA + \sqrt{H^4 SAT\iota}\big)$, with probability at least $1 - 2p$. Rescaling $p$ to $p/2$ finishes the proof. □

## Acknowledgements

We thank Nan Jiang, Sham M. Kakade, Greg Yang and Chicheng Zhang for valuable discussions. This work was supported in part by the DARPA program on Lifelong Learning Machines.

## References

[1] Shipra Agrawal and Randy Jia. Optimistic posterior sampling for reinforcement learning: worst-case regret bounds. In *NIPS*, pages 1184–1194, 2017.

[2] Mohammad Gheshlaghi Azar, Remi Munos, Mohammad Ghavamzadeh, and Hilbert J Kappen. Speedy q-learning. In *Proceedings of the 24th International Conference on Neural Information Processing Systems*, pages 2411–2419. Curran Associates Inc., 2011.

[3] Mohammad Gheshlaghi Azar, Rémi Munos, and Hilbert J. Kappen. On the sample complexity of reinforcement learning with a generative model. In *ICML*, 2012.

[4] Mohammad Gheshlaghi Azar, Rémi Munos, and Hilbert J. Kappen. Minimax PAC bounds on the sample complexity of reinforcement learning with a generative model. *Machine Learning*, 91(3):325–349, 2013.

[5] Mohammad Gheshlaghi Azar, Ian Osband, and Rémi Munos. Minimax regret bounds for reinforcement learning. In *ICML*, pages 263–272, 2017.

[6] Sébastien Bubeck, Nicolo Cesa-Bianchi, et al. Regret analysis of stochastic and nonstochastic multi-armed bandit problems. *Foundations and Trends® in Machine Learning*, 5(1):1–122, 2012.

[7] Marc Deisenroth and Carl E Rasmussen. Pilco: A model-based and data-efficient approach to policy search. In *Proceedings of the 28th International Conference on machine learning (ICML-11)*, pages 465–472, 2011.

[8] Eyal Even-Dar and Yishay Mansour. Learning rates for q-learning. *Journal of Machine Learning Research*, 5(Dec):1–25, 2003.

[9] Maryam Fazel, Rong Ge, Sham M Kakade, and Mehran Mesbahi. Global convergence of policy gradient methods for linearized control problems. *arXiv preprint arXiv:1801.05039*, 2018.

[10] Thomas Jaksch, Ronald Ortner, and Peter Auer. Near-optimal regret bounds for reinforcement learning. *Journal of Machine Learning Research*, 11:1563–1600, 2010.

[11] Nan Jiang, Akshay Krishnamurthy, Alekh Agarwal, John Langford, and Robert E Schapire. Contextual decision processes with low bellman rank are pac-learnable. *arXiv preprint arXiv:1610.09512*, 2016.

[12] Sham Kakade, Mengdi Wang, and Lin F Yang. Variance reduction methods for sublinear reinforcement learning. *ArXiv e-prints*, abs/1802.09184, April 2018.

[13] Sham M. Kakade. *On the sample complexity of reinforcement learning*. PhD thesis, University of London London, England, 2003.

[14] Michael Kearns and Satinder Singh. Near-optimal reinforcement learning in polynomial time. *Machine learning*, 49(2-3):209–232, 2002.

[15] Sven Koenig and Reid G Simmons. Complexity analysis of real-time reinforcement learning. In *AAAI*, pages 99–105, 1993.

[16] Tor Lattimore and Marcus Hutter. PAC bounds for discounted MDPs. In *ALT*, pages 320–334, 2012.

[17] Volodymyr Mnih, Koray Kavukcuoglu, David Silver, Alex Graves, Ioannis Antonoglou, Daan Wierstra, and Martin Riedmiller. Playing atari with deep reinforcement learning. *arXiv preprint arXiv:1312.5602*, 2013.

[18] Volodymyr Mnih, Adria Puigdomenech Badia, Mehdi Mirza, Alex Graves, Timothy Lillicrap, Tim Harley, David Silver, and Koray Kavukcuoglu. Asynchronous methods for deep reinforcement learning. In *International Conference on Machine Learning*, pages 1928–1937, 2016.

[19] Anusha Nagabandi, Gregory Kahn, Ronald S Fearing, and Sergey Levine. Neural network dynamics for model-based deep reinforcement learning with model-free fine-tuning. *arXiv preprint arXiv:1708.02596*, 2017.

[20] Ian Osband and Benjamin Van Roy. On lower bounds for regret in reinforcement learning. *ArXiv e-prints*, abs/1608.02732, April 2016.

[21] Vitchyr Pong, Shixiang Gu, Murtaza Dalal, and Sergey Levine. Temporal difference models: Model-free deep rl for model-based control. *arXiv preprint arXiv:1802.09081*, 2018.

[22] John Schulman, Sergey Levine, Pieter Abbeel, Michael Jordan, and Philipp Moritz. Trust region policy optimization. In *International Conference on Machine Learning*, pages 1889–1897, 2015.

[23] Aaron Sidford, Mengdi Wang, Xian Wu, and Yinyu Ye. Variance reduced value iteration and faster algorithms for solving markov decision processes. In *Proceedings of the Twenty-Ninth Annual ACM-SIAM Symposium on Discrete Algorithms*, pages 770–787. SIAM, 2018.

[24] David Silver, Aja Huang, Chris J Maddison, Arthur Guez, Laurent Sifre, George Van Den Driessche, Julian Schrittwieser, Ioannis Antonoglou, Veda Panneershelvam, Marc Lanctot, et al. Mastering the game of go with deep neural networks and tree search. *nature*, 529 (7587):484–489, 2016.

[25] Alexander L Strehl, Lihong Li, Eric Wiewiora, John Langford, and Michael L Littman. PAC model-free reinforcement learning. In *Proceedings of the 23rd international conference on Machine learning*, pages 881–888. ACM, 2006.

[26] Richard S Sutton and Andrew G Barto. *Reinforcement learning: An introduction*. MIT press Cambridge, 1998.

[27] Christopher John Cornish Hellaby Watkins. *Learning from delayed rewards*. PhD thesis, King's College, Cambridge, 1989.

# Appendix

## A  Explanation for Q-Learning with $\varepsilon$-Greedy

We recall a construction of a hard instance for Q-learning, known as a "combination lock," and tracing back at least to Koenig and Simmons [15]. In our context of our episodic MDP, this instance corresponds to the following MDP.

Consider a special state $s^\star \in \mathcal{S}$ where the adversary always picks $x_1 = s^\star$. For steps $h = 1, 2, \ldots, H/2$, there is one special action $a^\star \in \mathcal{A}$ where the distribution $\mathbb{P}_h(\cdot|s^\star, a^\star)$ is a singleton and always leads to a next state $x_{h+1} = s^\star$. For any other state $s \in \mathcal{S} \setminus \{s^\star\}$, or any other action $a \in \mathcal{A} \setminus \{a^\star\}$, the distribution $\mathbb{P}_h(\cdot|s, a)$ is uniform over $\mathcal{S} \setminus \{s^\star\}$. For steps $h = H/2 + 1, \ldots, H$, $\mathbb{P}_h(\cdot|s, a)$ is always a singleton and leads to the next state $x_{h+1} = s$. Finally, the reward function $r_h(s, a) = 0$ for all $s, a, h$, except when $s = s^\star$ and $h > H/2$, we have $r_H(s^\star, a^\star) = 1$. It is clear that the optimal policy gives reward $H/2$ (by always selecting action $a^\star$).

For this MDP, for the Q-learning algorithm (or its Sarsa variant) with zero initialization, unless the algorithm picks a path with prefix $(x_1, a_1, x_2, a_2, \ldots, x_{H/2}, a_{H/2}) = (s^\star, a^\star, \ldots, s^\star, a^\star)$, the reward value of the path is always zero and thus the algorithm will not change $Q_h(s, a)$ for any $s, a, h$. In other words, all Q values remain at zero until the first time $(s^\star, a^\star, \ldots, s^\star, a^\star)$ is visited. Unfortunately, this can happen with probability at most $A^{-H/2}$, and therefore the algorithm must suffer $H/2$ regret per round unless $K \geq \Omega(A^{H/2})$.

# B Proof of Lemma 4.1

In this section, we derive three important properties implied by our choice of the learning rate. Recall the notation from (3.1) and (4.2):

$$\alpha_t = \frac{H+1}{H+t}, \qquad \alpha_t^0 = \prod_{j=1}^t (1-\alpha_j), \qquad \alpha_t^i = \alpha_i \prod_{j=i+1}^t (1-\alpha_j) .$$

**Lemma 4.1.** *The following properties hold for $\alpha_t^i$:*

(a) $\frac{1}{\sqrt{t}} \leq \sum_{i=1}^t \frac{\alpha_t^i}{\sqrt{i}} \leq \frac{2}{\sqrt{t}}$ *for every $t \geq 1$.*

(b) $\max_{i \in [t]} \alpha_t^i \leq \frac{2H}{t}$ *and* $\sum_{i=1}^t (\alpha_t^i)^2 \leq \frac{2H}{t}$ *for every $t \geq 1$.*

(c) $\sum_{t=i}^\infty \alpha_t^i = 1 + \frac{1}{H}$ *for every $i \geq 1$.*

*Proof of Lemma 4.1.*

(a) The proof is by induction on $t$. For the base case $t = 1$ we have $\sum_{i=1}^t \frac{\alpha_t^i}{\sqrt{i}} = \alpha_1^1 = 1$ so the statement holds. For $t \geq 2$, by the relationship $\alpha_t^i = (1-\alpha_t)\alpha_{t-1}^i$ for $i = 1, 2, \ldots, t-1$ we have:

$$\sum_{i=1}^t \frac{\alpha_t^i}{\sqrt{i}} = \frac{\alpha_t}{\sqrt{t}} + (1-\alpha_t)\sum_{i=1}^{t-1} \frac{\alpha_{t-1}^i}{\sqrt{i}} .$$

On the one hand, by induction we have:

$$\frac{\alpha_t}{\sqrt{t}} + (1-\alpha_t)\sum_{i=1}^{t-1} \frac{\alpha_{t-1}^i}{\sqrt{i}} \geq \frac{\alpha_t}{\sqrt{t}} + \frac{1-\alpha_t}{\sqrt{t-1}} \geq \frac{\alpha_t}{\sqrt{t}} + \frac{1-\alpha_t}{\sqrt{t}} = \frac{1}{\sqrt{t}} .$$

On the other hand, by induction we have:

$$\frac{\alpha_t}{\sqrt{t}} + (1-\alpha_t)\sum_{i=1}^{t-1} \frac{\alpha_{t-1}^i}{\sqrt{i}} \leq \frac{\alpha_t}{\sqrt{t}} + \frac{2(1-\alpha_t)}{\sqrt{t-1}} = \frac{H+1}{\sqrt{t}(H+t)} + \frac{2\sqrt{t-1}}{H+t}$$

$$\leq \frac{H+1}{\sqrt{t}(H+t)} + \frac{2\sqrt{t}}{H+t} = \frac{2}{\sqrt{t}} + \frac{1}{\sqrt{t}} \cdot \frac{1-H}{t+H} \leq \frac{2}{\sqrt{t}} ,$$

where the final inequality holds because $H \geq 1$.

(b) We have:

$$\alpha_t^i = \frac{H+1}{i+H} \cdot \left(\frac{i}{i+1+H}\frac{i+1}{i+2+H}\cdots\frac{t-1}{t+H}\right)$$

$$= \frac{H+1}{t+H} \cdot \left(\frac{i}{i+H}\frac{i+1}{i+1+H}\cdots\frac{t-1}{t-1+H}\right) \leq \frac{H+1}{t+H} \leq \frac{2H}{t} .$$

Therefore, we have proved $\max_{i \in [t]} \alpha_t^i \leq 2H/t$. The second inequality, $\sum_{i=1}^t (\alpha_t^i)^2 \leq 2H/t$, follows directly since $\sum_{i=1}^t (\alpha_t^i)^2 \leq [\max_{i \in [t]} \alpha_t^i] \cdot \sum_{i=1}^t \alpha_t^i$ and $\sum_{i=1}^t \alpha_t^i = 1$.

(c) We first note the following identity, which holds for all positive integers $n$ and $k$ with $n \geq k$:

$$\frac{n}{k} = 1 + \frac{n-k}{n+1} + \frac{n-k}{n+1}\frac{n-k+1}{n+2} + \frac{n-k}{n+1}\frac{n-k+1}{n+2}\frac{n-k+2}{n+3} + \cdots . \qquad \text{(B.1)}$$

To verify (B.1), we write the terms of its right-hand side as $x_0 = 1, x_1 = \frac{n-k}{n+1}, \ldots$. It is easy to verify by induction that $\frac{n}{k} - \sum_{i=0}^t x_i = \frac{n-k}{k} \prod_{i=1}^t \frac{n-k+i}{n+i}$. This means $\lim_{t \to \infty} \frac{n}{k} - \sum_{i=0}^t x_i = 0$ and this proves that (B.1) holds. Now, using (B.1) with $n = i + H$ and $k = H$, we have:

$$\sum_{t=i}^{\infty} \alpha_t^i = \frac{H+1}{i+H} \cdot \left(1 + \frac{i}{i+1+H} + \frac{i}{i+1+H}\frac{i+1}{i+2+H} + \cdots\right) = \frac{H+1}{i+H} \cdot \frac{i+H}{H} = \frac{H+1}{H} \ .$$

□

## C  Proof for Q-learning with UCB-Bernstein

In this section, we prove Theorem 2.

**Notation.** In addition to the notation of Section 4, we define a variance operator $\mathbb{V}_h$:

$$[\mathbb{V}_h V_{h+1}](x,a) := \text{Var}_{x' \sim \mathbb{P}_h(\cdot|x,a)}(V_{h+1}(x')) = \mathbb{E}_{x' \sim \mathbb{P}_h(\cdot|x,a)} \left[V_{h+1}(x') - [\mathbb{P}_h V_{h+1}](x,a)\right]^2$$

We also consider an empirical version of variance that can be computed by the algorithm: when $(x,a)$ was taken at step $h$ for $t$ times at $k_1, \cdots, k_t$ episodes respectively:

$$W_t(x, a, h) := \frac{1}{t} \sum_{i=1}^{t} \left[ V_{h+1}^{k_i}(x_{h+1}^{k_i}) - \frac{1}{t} \sum_{j=1}^{t} V_{h+1}^{k_j}(x_{h+1}^{k_j}) \right]^2 \ . \tag{C.1}$$

In this section, we choose two constants $c_1, c_2 > 0$ and define

$$\beta_t(x, a, h) := \min \left\{ c_1 \left( \sqrt{\frac{H}{t} \cdot (W_t(x,a,h) + H)\iota} + \frac{\sqrt{H^7 SA \cdot \iota}}{t} \right), c_2 \sqrt{\frac{H^3 \iota}{t}} \right\} \ , \tag{C.2}$$

and accordingly,

$$b_1(x, a, h) := \frac{\beta_1(x, a, h)}{2} \qquad b_t(x, a, h) := \frac{\beta_t(x, a, h) - (1 - \alpha_t)\beta_{t-1}(x, a, h)}{2\alpha_t} \ . \tag{C.3}$$

It is easy to verify that $\beta_t = 2 \sum_{i=1}^{t} \alpha_t^i b_i$ for every $t \geq 1$. We include in Algorithm 2 the efficient implementation for calculating $b_t(x, a, h)$ in $O(1)$ time per time step. Now we restate Theorem 2.

**Theorem 2** (Bernstein, restated). *There exist absolute constants $c_1, c_2 > 0$ such that, for any $p \in (0,1)$, if we choose $b_t$ according to (C.3), then with probability $1 - p$, the total regret of Q-learning with UCB-Bernstein (see Algorithm 2) is at most $O(\sqrt{H^3 SAT \iota} + \sqrt{H^9 S^3 A^3} \cdot \iota^2)$.*

### C.1  Proof

We first note that the following recursion, obtained in the proof for the Hoeffding case (see Lemma 4.2), still holds here:

**Lemma C.1** (recursion on Q). *For any $(x, a, h) \in \mathcal{S} \times \mathcal{A} \times [H]$ and episode $k \in [K]$, let $t = N_h^k(x, a)$ and suppose $(x, a)$ was previously taken at step $h$ of episodes $k_1, \ldots, k_t < k$, then*

$$(Q_h^k - Q_h^\star)(x, a) = \alpha_t^0 (H - Q_h^\star(x,a))$$
$$+ \sum_{i=1}^{t} \alpha_t^i \left[ (V_{h+1}^{k_i} - V_{h+1}^\star)(x_{h+1}^{k_i}) + [(\hat{\mathbb{P}}_h^{k_i} - \mathbb{P}_h)V_{h+1}^\star](x, a) + b_i(x, a, h) \right] \ .$$

Parallel to the Hoeffding case, we aim at proving an equivalent version of Lemma 4.3 that shows that $Q^k - Q^\star$ is (1) nonnegative and (2) bounded from above. However, unlike the Hoeffding case, this new proof becomes very delicate.

**Algorithm 2** Q-learning with UCB-Bernstein

1: **for all** $(x, a, h) \in \mathcal{S} \times \mathcal{A} \times [H]$ **do**
2:     $Q_h(x, a) \leftarrow H$;   $N_h(x, a) \leftarrow 0$;   $\mu_h(x, a) \leftarrow 0$;   $\sigma_h(x, a) \leftarrow 0$;   $\beta_0(x, a, h) \leftarrow 0$.
3: **for** episode $k = 1, \ldots, K$ **do**
4:     receive $x_1$.
5:     **for** step $h = 1, \ldots, H$ **do**
6:        Take action $a_h \leftarrow \text{argmax}_{a'} Q_h(x_h, a')$, and observe $x_{h+1}$.
7:        $t = N_h(x_h, a_h) \leftarrow N_h(x_h, a_h) + 1$.
8:        $\mu_h(x_h, a_h) \leftarrow \mu_h(x_h, a_h) + V_{h+1}(x_{h+1})$.
9:        $\sigma_h(x_h, a_h) \leftarrow \sigma_h(x_h, a_h) + \big(V_{h+1}(x_{h+1})\big)^2$.
10:       $\beta_t(x_h, a_h, h) \leftarrow \min\left\{ c_1 \Big( \sqrt{\frac{H}{t} \frac{\sigma_h(x_h, a_h) - (\mu_h(x_h, a_h))^2}{t} + H)\iota} + \frac{\sqrt{H^7 SA \cdot \iota}}{t} \Big), c_2 \sqrt{\frac{H^3 \iota}{t}} \right\}$.
11:       $b_t \leftarrow \frac{\beta_t(x_h, a_h, h) - (1 - \alpha_t)\beta_{t-1}(x_h, a_h, h)}{2\alpha_t}$.
12:       $Q_h(x_h, a_h) \leftarrow (1 - \alpha_t) Q_h(x_h, a_h) + \alpha_t [r_h(x_h, a_h) + V_{h+1}(x_{h+1}) + b_t]$.
13:       $V_h(x_h) \leftarrow \min\{H, \max_{a' \in \mathcal{A}} Q_h(x_h, a')\}$.

We first provide a *coarse* upper bound on $Q^k - Q^\star$ that does not assert whether $Q^k - Q^\star$ is nonnegative or not. This coarse upper bound only makes use of the fact that $\beta_t$ is at most $O(\sqrt{H^3 \iota / t})$, which was precisely how we have chosen $\beta_t$ in the Hoeffding case and in Lemma 4.3.

**Lemma C.2** (coarse bound on $Q^k - Q^\star$). *There exists absolute constant $c_2 > 0$ such that, if $\beta_t(x, a, h) \leq c_2 \sqrt{\frac{H^3 \iota}{t}}$ in (C.2), then, with probability at least $1 - p$, the following holds*

$\forall (x, a, h, k) \in \mathcal{S} \times \mathcal{A} \times [H] \times [K]$:

$$(V_h^k - V_h^\star)(x_h^k) \leq \alpha_t^0 H + \sum_{i=1}^t \alpha_t^i (V_{h+1}^{k_i} - V_{h+1}^\star)(x_{h+1}^{k_i}) + 4c_2 \sqrt{\frac{H^3 \iota}{t}}, \quad (C.4)$$

*where $t = N_h^k(x, a)$ and $k_1, \ldots, k_t < k$ are the episodes in which $(x, a)$ was taken at step $h$.*

*Proof of Lemma C.2.* The result follows from Lemma C.1 and the proof of Lemma 4.3. □

In order to apply the Bernstein concentration inequality to the recursive formula in Lemma C.1, we need to estimate the variance of $V^\star$. Unfortunately, $V^\star$ is unknown as its variance. At the $k$th episode, we are only able to compute the "empirical" version of the variance using $V^k$, which is $W_t$ as defined in (C.1).

Our next lemma shows that, if $Q^{k'} - Q^\star$ is nonnegative for all episodes $k' < k$, the variance of $V^\star$ (i.e., $\mathbb{V}_h V_{h+1}^\star(x, a)$) and the "empirical" variance of $V^k$ are sufficiently close.

**Lemma C.3.** *There exists an absolute constant $c > 0$ such that for any $p \in (0, 1)$ and $k \in [K]$, with probability at least $1 - p/K$, if*

*(C.4) in Lemma C.2 holds and $(Q_h^{k'} - Q_h^\star)(x, a) \geq 0$ for all $k' < k$,*

*then for all $(x, a, h) \in \mathcal{S} \times \mathcal{A} \times [H]$:*

$$\big|\mathbb{V}_h V_{h+1}^\star(x, a) - W_t(x, a, h)\big| \leq c \Big(\frac{SA\sqrt{H^7 \iota}}{t} + \sqrt{\frac{H^7 SA \iota}{t}}\Big), \quad \text{where} \quad t = N_h^k(x, a).$$

*Proof of Lemma C.3.* For each fixed $(x, a, h) \in \mathcal{S} \times \mathcal{A} \times [H]$, let us denote $k_0 = 0$, and:

$$k_i = \min\big(\{k \in [K] \mid k > k_{i-1} \wedge (x_h^k, a_h^k) = (x, a)\} \cup \{K + 1\}\big).$$

That is, $k_i$ is the episode if which $(x, a)$ was taken at step $h$ for the $i$th time, and it is clearly a stopping time. Let $\mathcal{F}_i$ be the $\sigma$-field generated by all the random variables until episode $k_i$, step $h$. We also denote $t = N_h^k(x, a)$.

To bridge the gap between $\mathbb{V}_h V_{h+1}^\star(x, a)$ and $W_t(x, a, h)$, we consider following four quantities:

$$[\mathbb{V}_h V_{h+1}^\star](x, a) = \mathbb{E}_{x' \sim \mathbb{P}_h(\cdot | x, a)} \left[ V_{h+1}^\star(x') - [\mathbb{P}_h V_{h+1}^\star](x, a) \right]^2 \qquad =: P_1$$

$$\frac{1}{t} \sum_{i=1}^{t} \left[ V_{h+1}^\star(x_{h+1}^{k_i}) - [\mathbb{P}_h V_{h+1}^\star](x, a) \right]^2 \qquad =: P_2$$

$$\frac{1}{t} \sum_{i=1}^{t} \left[ V_{h+1}^\star(x_{h+1}^{k_i}) - \tfrac{1}{t} \sum_{j=1}^{t} V_{h+1}^\star(x_{h+1}^{k_j}) \right]^2 \qquad =: P_3$$

$$W_t(x, a, h) = \frac{1}{t} \sum_{i=1}^{t} \left[ V_{h+1}^{k_i}(x_{h+1}^{k_i}) - \tfrac{1}{t} \sum_{j=1}^{t} V_{h+1}^{k_j}(x_{h+1}^{k_j}) \right]^2 \qquad =: P_4 \ .$$

We shall bound the difference $|P_1 - P_4|$ by $|P_1 - P_2| + |P_2 - P_3| + |P_3 - P_4|$ via the triangle inequality.

**Bounding $|P_1 - P_2|$:** We notice that for any fixed $\tau \in [k]$, by the Azuma-Hoeffding inequality, there exists a sufficiently large constant $c > 0$ such that, with probability at least $1 - p/(2SAT)$:

$$\left| \frac{1}{\tau} \sum_{i=1}^{\tau} \mathbb{I}[k_i \leq k] \cdot \left[ \left( V_{h+1}^\star(x_{h+1}^{k_i}) - [\mathbb{P}_h V_{h+1}^\star](x, a) \right)^2 - [\mathbb{V}_h V_{h+1}](x, a) \right] \right| \leq cH^2 \sqrt{\iota/\tau} \ , \qquad \text{(C.5)}$$

since LHS is a martingale sequence with respect to the filtration $\{\mathcal{F}_i\}$. Because Eq. (C.5) holds for all fixed $\tau \in [k]$ uniformly, it also holds for $\tau = t = N_h^k(x, a) \leq k$ which is a random variable. Also note $\mathbb{I}[k_i \leq k] = 1$ for all $i \leq N_h^k(x, a)$. Therefore, we can conclude $|P_1 - P_2| \leq cH^2 \sqrt{\iota/t}$.

**Bounding $|P_2 - P_3|$:** We calculate

$$|P_2 - P_3| \leq \frac{2H}{t} \sum_{i=1}^{t} \left| [\mathbb{P}_h V_{h+1}^\star](x, a) - \tfrac{1}{t} \sum_{j=1}^{t} V_{h+1}^\star(x_{h+1}^{k_j}) \right| = 2H \left| [\mathbb{P}_h V_{h+1}^\star](x, a) - \tfrac{1}{t} \sum_{j=1}^{t} V_{h+1}^\star(x_{h+1}^{k_j}) \right| \ .$$

Again, for any fixed $\tau \in [k]$, by the Azuma-Hoeffding inequality, with probability $1 - p/(2SAT)$:

$$\left| \frac{1}{\tau} \sum_{i=1}^{\tau} \mathbb{I}[k_i \leq k] \cdot \left[ V_{h+1}^\star(x_{h+1}^{k_i}) - \mathbb{P}_h V_{h+1}^\star(x, a) \right] \right| \leq cH \sqrt{\iota/\tau} \ . \qquad \text{(C.6)}$$

By the same argument as above, we also know that Eq. (C.5) holds for the random variable $\tau = t = N_h^k(x, a) \leq k$, which implies $|P_2 - P_3| \leq 2cH^2 \sqrt{\iota/t}$.

**Bounding $|P_3 - P_4|$:** We calculate that

$$|P_3 - P_4| \leq \frac{2H}{t} \sum_{i=1}^{t} \left| V_{h+1}^{k_i}(x_{h+1}^{k_i}) - V_{h+1}^\star(x_{h+1}^{k_i}) - \tfrac{1}{t} \sum_{j=1}^{t} \left( V_{h+1}^{k_j}(x_{h+1}^{k_j}) - V_{h+1}^\star(x_{h+1}^{k_j}) \right) \right|$$

$$\leq \frac{4H}{t} \sum_{i=1}^{t} \left| V_{h+1}^{k_i}(x_{h+1}^{k_i}) - V_{h+1}^\star(x_{h+1}^{k_i}) \right| \leq \frac{4H}{t} \sum_{i=1}^{t} \left( V_{h+1}^{k_i}(x_{h+1}^{k_i}) - V_{h+1}^\star(x_{h+1}^{k_i}) \right) \ ,$$

where the last inequality uses $V_{h+1}^{k'}(x) \geq V_{h+1}^\star(x)$ for all $x \in \mathcal{S}$ and $k' < k$, which follows from our assumption $(Q_{h+1}^{k'} - Q_{h+1}^\star)(x, a) \geq 0$ for all $k' < k$.

We apply Lemma C.7 (see Section C.3 later) with a weight vector $w$ such that $w_{k_i} = \frac{1}{t}$ for all $i \in [t]$, but $w_{k'} = 0$ for all $k' \notin \{k_1, \ldots, k_t\}$ (so $\|w\|_1 = 1$ and $\|w\|_\infty = 1/t$). This tells us that

$$|P_3 - P_4| \leq \frac{4H}{t} \sum_{i=1}^{t} \left(V_{h+1}^{k_i}(x_{h+1}^{k_i}) - V_{h+1}^{\star}(x_{h+1}^{k_i})\right) \leq O\left(\frac{SA\sqrt{H^7 \iota}}{t} + \sqrt{\frac{H^7 SA\iota}{t}}\right) .$$

Finally, by the triangle inequality $\left|[\mathbb{V}_h V_{h+1}^\star - W_h^k](x,a)\right| \leq |P_1 - P_2| + |P_2 - P_3| + |P_3 - P_4|$, and a union bound over $(x,a,h) \in \mathcal{S} \times \mathcal{A} \times [H]$, we finish the proof. $\square$

Now, equipped with Lemma C.2 and Lemma C.3, we can use induction and an Azuma-Bernstein concentration argument to prove that $Q^k - Q^\star$ is nonnegative and upper bounded by $\beta$. This gives an analog of Lemma 4.3 that we state here.

**Lemma C.4** (fine bound on $Q^k - Q^\star$). *For every $p \in (0,1)$, there exists an absolute constant $c_1, c_2 > 0$ such that, under the choice of $\beta_t(x,a,h)$ in (C.2), with probability at least $1 - 2p$, the following holds simultaneously for all $(x, a, h, k) \in \mathcal{S} \times \mathcal{A} \times [H] \times [K]$:*

$$0 \leq (Q_h^k - Q_h^\star)(x,a) \leq \alpha_t^0 H + \sum_{i=1}^{t} \alpha_t^i (V_{h+1}^{k_i} - V_{h+1}^{\star})(x_{h+1}^{k_i}) + \beta_t , \quad (C.7)$$

*where $t = N_h^k(x,a)$ and $k_1, \ldots, k_t < k$ are the episodes in which $(x,a)$ was taken at step $h$.*

*Proof of Lemma C.4.* We first choose $c_2 > 0$ large enough so that Lemma C.2 holds with probability at least $1 - p$.

For each fixed $(x, a, h) \in \mathcal{S} \times \mathcal{A} \times [H]$, let us denote $k_0 = 0$, and:

$$k_i = \min\left(\{k \in [K] \mid k > k_{i-1} \wedge (x_h^k, a_h^k) = (x,a)\} \cup \{K+1\}\right) .$$

By the Azuma-Bernstein inequality, with probability at least $1 - p/(SAT)$, we have for all $\tau \in [K]$:

$$\left|\sum_{i=1}^{\tau} \alpha_\tau^i \mathbb{I}[k_i \leq K] \cdot [(\hat{\mathbb{P}}_h^{k_i} - \mathbb{P}_h) V_{h+1}^\star](x,a)\right| \leq O(1) \cdot \left[\sqrt{\sum_{i=1}^{\tau} (\alpha_\tau^i)^2 [\mathbb{V}_h V_{h+1}^\star](x,a) \iota} + [\max_{i \in [\tau]} \alpha_\tau^i] H \iota\right]$$

$$\leq O(1) \cdot \left[\sqrt{\frac{H}{\tau} [\mathbb{V}_h V_{h+1}^\star](x,a) \iota} + \frac{H^2}{\tau} \iota\right] , \quad (C.8)$$

where the last inequality is by Lemma 4.1.b. Since the inequality (C.8) holds for all fixed $\tau \in [K]$ uniformly, it also holds for the random variable $\tau = t = N_h^k(x,a) \leq K$. By a union bound, with probability at least $1 - p$, we have that for all $(x, a, h, k) \in \mathcal{S} \times \mathcal{A} \times [H] \times [K]$

$$\left|\sum_{i=1}^{t} \alpha_t^i \mathbb{I}[k_i \leq K] \cdot [(\hat{\mathbb{P}}_h^{k_i} - \mathbb{P}_h) V_{h+1}^\star](x,a)\right| \leq O(1) \cdot \left[\sqrt{\frac{H}{t} [\mathbb{V}_h V_{h+1}^\star](x,a) \iota} + \frac{H^2}{t} \iota\right] , \quad (C.9)$$

where $t = N_h^k(x,a)$ and $k_1, \ldots, k_t < k$ are the episodes in which $(x,a)$ was taken at step $h$.

We are now ready to prove (C.7). We do so by induction over $k \in [K]$. Clearly, the statement is true for $k = 1$, so in the rest of the proof we assume (C.7) holds for all $k' < k$. We denote by $k_1, k_2, \ldots, k_t < k$ all indices of previous episodes where $(x,a)$ is taken at step $h$. By Lemma C.3, with probability $1 - p/K$, we have for all $(x, a, h) \in \mathcal{S} \times \mathcal{A} \times [H]$:

$$\left|[\mathbb{V}_h V_{h+1}^\star](x,a) - W_t(x,a,h)\right| \leq O\left(\sqrt{\frac{SAH^7 \iota}{t}} + \frac{SA\sqrt{H^7 \iota}}{t}\right) .$$

Therefore, putting this into (C.9), we have

$$\left|\sum_{i=1}^{t} \alpha_t^i [(\hat{\mathbb{P}}_h^{k_i} - \mathbb{P}_h) V_{h+1}^\star](x,a)\right| \overset{①}{\leq} O(1) \cdot \left[\sqrt{\frac{H}{t}(W_t(x,a,h) + H)\iota} + \frac{\sqrt{H^7 SA} \cdot \iota}{t}\right] \overset{②}{\leq} \frac{\beta_t}{2} ,$$

where inequality ① uses $\sqrt{\frac{H^7 SA\iota}{t}} \leq H + \frac{H^6 SA\iota}{t}$, and inequality ② is due to our choice of $\beta_t$ in (C.2) and the sufficiently large choice of $c_1 > 0$.

Finally, applying the above inequality to Lemma C.1, we have for all $(x,a,h) \in \mathcal{S} \times \mathcal{A} \times [H]$

$$0 \leq (Q_h^k - Q_h^\star)(x,a) - \alpha_t^0(H - Q_h^\star(x,a)) - \sum_{i=1}^{t} \alpha_t^i [(V_{h+1}^{k_i} - V_{h+1}^\star)(x_{h+1}^{k_i})] \leq \beta_t . \quad (\text{C.10})$$

This proves that (C.7) holds for $k$ with probability at least $1 - p/K$. By induction, we know (C.7) holds for all $k \in [K]$ with probability at least $1 - p$. Combining this with the $1 - p$ probability event for (C.9), we finish the proof that Lemma C.4 holds with probability at least $1 - 2p$. □

As mentioned in Section 3, the key reason why a Bernstein approach can improve by a factor of $\sqrt{H}$ is that, although the value function at each step is at most $H$, the "total variance of the value function" for an entire episode is at most $O(H^2)$. Or more simply, the total variance for all steps is at most $O(HT)$. This is captured directly in the following lemma.

**Lemma C.5.** *There exists an absolute constant $c$, such that with probability at least $1 - p$:*

$$\sum_{k=1}^{K} \sum_{h=1}^{H} \mathbb{V}_h V_{h+1}^{\pi_k}(x_h^k, a_h^k) \leq c(HT + H^3 \iota) .$$

*Proof of Lemma C.5.* First, we note for any fixed policy $\pi$ and initial state $x_1$, suppose $(x_2, \cdots, x_h)$ is a sequence generated by following policy $\pi$ starting at $x_1$, then

$$H^2 \geq \mathbb{E}\left[\left(\sum_{h=1}^{H} r(x_h, \pi(x_h))\right) - V_1^\pi(x_1)\right]^2$$

$$\overset{①}{=} \mathbb{E}\left[\sum_{h=1}^{H} [r(x_h, \pi(x_h)) + V_{h+1}^\pi(x_{h+1}) - V_h^\pi(x_h)]\right]^2$$

$$\overset{②}{=} \mathbb{E}\sum_{h=1}^{H} \left[r(x_h, \pi(x_h)) + V_{h+1}^\pi(x_{h+1}) - V_h^\pi(x_h)\right]^2 = \mathbb{E}\sum_{h=1}^{H} \mathbb{V}_h V_{h+1}^\pi(x_h, \pi(x_h)) ,$$

where equality ① is because $V_{H+1}^\pi = 0$, and equality ② uses the independence due to the Markov property. Therefore, letting $\mathcal{F}_{k-1}$ be the $\sigma$-field generated by all the random variables over the first $k-1$ episodes, at the $k$th episode we have:

$$\mathbb{E}\left[X_k \,\Big|\, \mathcal{F}_{k-1}\right] \leq H^2 \quad \text{where} \quad X_k := \sum_{h=1}^{H} \mathbb{V}_h V_{h+1}^{\pi_k}(x_h^k, \pi_k(x_h^k)) .$$

Also, note that $|X_k| \leq H^3$ and $\mathrm{Var}[X_k \,|\, \mathcal{F}_{k-1}] \leq H^3 \mathbb{E}[X_k \,|\, \mathcal{F}_{k-1}] \leq H^5$. Therefore, by an Azuma-Bernstein inequality on $X_1 + \cdots + X_K$ with respect to filtration $\{\mathcal{F}_k\}_{k \geq 0}$, we have with probability at least $1 - p$,

$$\sum_{k=1}^{K} \sum_{h=1}^{H} \mathbb{V}_h V_{h+1}^{\pi_k}(x_h^k, a_h^k) \leq \sum_{k=1}^{K} \mathbb{E}[X_k \,|\, \mathcal{F}_{k-1}] + O(\sqrt{H^5 K \iota} + H^3 \iota) \leq O(HT + H^3 \iota) ,$$

where the last step is by $ab \leq a^2 + b^2$. □

Our last lemma shows that the "empirical" variance of $V^k$ (i.e., $W_t(x,a,h)$) is also upper bounded by the variance $\mathbb{V}_h V_{h+1}^{\pi_k}(x,a)$ (which appeared in Lemma C.5) plus some small terms.

**Lemma C.6.** *There exist absolute constants $c_1, c_2, c > 0$ such that, letting $(x, a) = (x_h^k, a_h^k)$ and $t = n_h^k = N_h^k(x, a)$, we have that for all $(k, h) \in [K] \times [H]$, with probability at least $1 - 4p$,*

$$W_t(x, a, h) \leq \mathbb{V}_h V_{h+1}^{\pi_k}(x, a) + 2H(\delta_{h+1}^k + \xi_{h+1}^k) + c\left(\frac{SA\sqrt{H^7\iota}}{t} + \sqrt{\frac{SAH^7\iota}{t}}\right),$$

*where $\xi_{h+1}^k := [(\mathbb{P}_h - \hat{\mathbb{P}}_h^k)(V_{h+1}^\star - V_{h+1}^k)](x_h^k, a_h^k)$ and $\delta_{h+1}^k := (V_{h+1}^\star - V_{h+1}^k)(x_{h+1}^k)$.*

*Proof of Lemma C.6.* We first assume that Lemma C.4 holds (which happens with probability at least $1-2p$) and Lemma C.2 holds (which happens with probability at least $1-p$). As a consequence, with probability at least $1 - p$, Lemma C.3 also holds for all $k \in [K]$. By the triangle inequality, we have:

$$W_t(x, a, h) - \mathbb{V}_h V_{h+1}^{\pi_k}(x, a) \leq \left|[\mathbb{V}_h V_{h+1}^\star - W_t(x, a, h)]\right| + \left|[\mathbb{V}_h V_{h+1}^\star - \mathbb{V}_h V_{h+1}^{\pi_k}](x, a)\right|,$$

where the first term on the right-hand side is upper bounded by Lemma C.3. For the second term:

$$\left|[\mathbb{V}_h V_{h+1}^\star - \mathbb{V}_h V_{h+1}^{\pi_k}](x, a)\right| \leq 2H[\mathbb{P}_h(V_{h+1}^\star - V_{h+1}^{\pi_k})](x_h^k, a_h^k) = 2H(\xi_{h+1}^k + \delta_{h+1}^k). \quad \square$$

## C.2 Proof of Theorem 2

We are now ready to prove Theorem 2. Again, the proof decomposes the regret in a recursive form, and carefully controls the error propagation via repeated usage of Lemma C.4 and Lemma C.6.

*Proof of Theorem 2.* We first assume that Lemma C.5 holds (which happens with probability at least $1 - 4p$) and Lemma C.6 holds (which happens with probability at least $1 - p$).

By the same argument as in the proof of Theorem 1 (in particular, inequality (4.7)) we have:

$$\sum_{k=1}^K \delta_h^k \leq \left(1 + \frac{1}{H}\right) \sum_{k=1}^K \delta_{h+1}^k + SAH + \sum_{k=1}^K (\beta_{n_h^k}(x_h^k, a_h^k, h) + \xi_{h+1}^k),$$

where $\xi_{h+1}^k := [(\mathbb{P}_h - \hat{\mathbb{P}}_h^k)(V_{h+1}^\star - V_{h+1}^k)](x_h^k, a_h^k)$ and $\delta_{h+1}^k := (V_{h+1}^\star - V_{h+1}^k)(x_{h+1}^k)$. As a result, for any $h \in H$, by recursing the above formula for $h, h+1, \ldots, H$, we have:

$$\sum_{k=1}^K \delta_h^k \leq SAH^2 + \sum_{h'=h}^H \sum_{k=1}^K (\beta_{n_{h'}^k}(x_{h'}^k, a_{h'}^k, h') + \xi_{h'+1}^k) \quad (C.11)$$

By the Azuma-Hoeffding inequality, with probability $1 - p$, we have:

$$\forall h \in [H]: \left|\sum_{h'=h}^H \sum_{k=1}^K \xi_{h'+1}^k\right| = \left|\sum_{h'=h}^H \sum_{k=1}^K [(\mathbb{P}_{h'} - \hat{\mathbb{P}}_{h'}^k)(V_{h'+1}^\star - V_{h'+1}^k)](x_{h'}^k, a_{h'}^k)\right| \leq O(H\sqrt{T\iota}). \quad (C.12)$$

Also, recall $\beta_t(x, a, h) \leq c\sqrt{H^3\iota/t}$ so $\sum_{k=1}^K \beta_{n_h^k} \leq O(\sqrt{H^2SAT\iota})$ according to (4.8). Putting these into (C.11), we derive that $\sum_{k=1}^K \delta_h^k \leq O(SAH^2 + \sqrt{H^4SAT\iota})$. Note when $T \geq \sqrt{H^4SAT\iota}$, we have $\sqrt{H^4SAT\iota} \geq H^2SA$; when $T \leq \sqrt{H^4SAT\iota}$, we have $\sum_{k=1}^K \delta_h^k \leq HK = T \leq \sqrt{H^4SAT\iota}$. Therefore, we can simply write

$$\sum_{k=1}^K \delta_h^k \leq O(\sqrt{H^4SAT\iota}). \quad (C.13)$$

By our choice of $\beta_t$, we have:

$$\sum_{k=1}^K \sum_{h=1}^H \beta_{n_h^k} \leq O(1) \cdot \sum_{k=1}^K \sum_{h=1}^H \left[\sqrt{\frac{H}{n_h^k} \cdot (W_{n_h^k}(x, a, h) + H)} + \frac{\sqrt{H^7SA} \cdot \iota}{n_h^k}\right] \quad (C.14)$$

The summation of the second term in (C.14) is upper bounded by

$$\sum_{k=1}^{K}\sum_{h=1}^{H}\frac{\sqrt{H^7 SA}\cdot\iota}{n_h^k} \leq \sqrt{H^9 S^3 A^3 \iota^4}\ ,$$

because $1+\frac{1}{2}+\frac{1}{3}+\cdots \leq \iota$. The summation of the first term in (C.14) can be upper bounded by

$$\sum_{k=1}^{K}\sum_{h=1}^{H}\sqrt{\frac{H}{n_h^k}\cdot(W_{n_h^k}(x,a,h)+H)} \leq \sqrt{\left(\sum_{k=1}^{K}\sum_{h=1}^{H}(W_{n_h^k}(x,a,h)+H)\right)\left(\sum_{k=1}^{K}\sum_{h=1}^{H}\frac{H}{n_h^k}\right)}$$
$$\leq \sqrt{\sum_{k=1}^{K}\sum_{h=1}^{H}W_{n_h^k}(x,a,h)}\cdot\sqrt{H^2 SA\iota}+\sqrt{H^3 SAT\iota}\ . \quad (C.15)$$

We calculate

$$\sum_{k=1}^{K}\sum_{h=1}^{H}W_{n_h^k}(x,a,h) \stackrel{①}{\leq} \sum_{k=1}^{K}\sum_{h=1}^{H}\left[\mathbb{V}_h V_{h+1}^{\pi_k}(x_h^k,a_h^k)+2H(\delta_{h+1}^k+\xi_{h+1}^k)+O\left(\frac{SA\sqrt{H^7\iota}}{n_h^k}+\sqrt{\frac{SAH^7\iota}{n_h^k}}\right)\right]$$
$$\stackrel{②}{\leq} \sum_{k=1}^{K}\sum_{h=1}^{H}\left[\mathbb{V}_h V_{h+1}^{\pi_k}(x_h^k,a_h^k)+2H(\delta_{h+1}^k+\xi_{h+1}^k)\right]+O\left(S^2 A^2\sqrt{H^9\iota^3}+SA\sqrt{H^8 T\iota}\right)$$
$$\stackrel{③}{\leq} 2H\sum_{k=1}^{K}\sum_{h=1}^{H}(\delta_{h+1}^k+\xi_{h+1}^k)+O\left(HT+H^3\iota+S^2 A^2\sqrt{H^9\iota^3}+SA\sqrt{H^8 T\iota}\right)$$
$$\stackrel{④}{\leq} O\left(\sqrt{H^8 SAT\iota}+HT+H^3\iota+S^2 A^2\sqrt{H^9\iota^3}+SA\sqrt{H^8 T\iota}\right)$$
$$\leq O\left(HT+S^2 A^2 H^7\iota+S^2 A^2\sqrt{H^9\iota^3}\right)\ . \quad (C.16)$$

Here, inequality ① uses Lemma C.6; inequality ② uses $\sum_{k=1}^{K}(n_h^k)^{-1}\leq SA\iota$ and $\sum_{k=1}^{K}(\sqrt{n_h^k})^{-1/2}\leq O(\sqrt{KSA})$; inequality ③ uses Lemma C.5; and inequality ④ uses (C.12) and (C.13).

Putting (C.16) and (C.15) back to (C.14), we have

$$\sum_{k=1}^{K}\sum_{h=1}^{H}\beta_{n_h^k} \leq O\left(\sqrt{H^3 SAT\iota}+\sqrt{S^3 A^3 H^9\iota^4}\right)\ . \quad (C.17)$$

Finally, putting this and (C.12) back to (C.11), we finish the proof that with probability at least $1-6p$, for every $h\in[H]$

$$\sum_{k=1}^{K}\delta_h^k \leq O\left(\sqrt{H^3 SAT\iota}+\sqrt{S^3 A^3 H^9\iota^4}\right)\ .$$

Since we also have $\text{Regret}(K)\leq \sum_{k=1}^{K}\delta_1^k$ as in the proof of Theorem 1, rescaling $p$ to $p/6$ finishes the proof. □

### C.3 Proof of Auxiliary Lemma

The next lemma shows how the weighted sum over $(V_h^k-V_h^\star)(x_h^k)$ is upper bounded by the infinity norm and the one-norm of the weights $w$. This lemma provides the key to prove Lemma C.3.

**Lemma C.7.** *Suppose (C.4) in Lemma C.2 holds. For any $h\in[H]$, let $\phi_h^k:=(V_h^k-V_h^\star)(x_h^k)$, and*

*letting $w = (w_1, \ldots, w_k)$ be a nonnegative weight vector, we have:*

$$\sum_{k=1}^{K} w_k \phi_h^k \leq O\big(SA\|w\|_\infty \sqrt{H^5 \iota} + \sqrt{SA\|w\|_1 \|w\|_\infty H^5 \iota}\big) \;,$$

*where $\phi_h^k := (V_h^k - V_h^\star)(x_h^k)$.*

*Proof of Lemma C.7.* For any fixed $(k, h) \in [K] \times [H]$, let $t = N_h^k(x_h^k, a_h^k)$, and suppose $(x_h^k, a_h^k)$ was previously taken at step $h$ of episodes $k_1, \ldots, k_t < k$. We then have, for some absolute constant $c$:

$$\phi_h^k = (V_h^k - V_h^\star)(x_h^k) \overset{①}{\leq} (Q_h^k - Q_h^\star)(x_h^k, a_h^k) \overset{②}{\leq} \alpha_t^0 H + \sum_{i=1}^{t} \alpha_t^i \phi_{h+1}^{k_i} + O\Big(\sqrt{\frac{H^3 \iota}{t}}\Big) \;. \quad (C.18)$$

Here, inequality ① holds from $V_h^k(x_h^k) \leq \max_{a' \in \mathcal{A}} Q_h^k(x_h^k, a') = Q_h^k(x_h^k, a_h^k)$ and the Bellman optimality equation $V_h^\star(x_h^k) = \max_{a' \in \mathcal{A}} Q_h^\star(x_h^k, a') \geq Q_h^\star(x_h^k, a_h^k)$. Inequality ② holds by the assumption that (C.4) in Lemma C.2 holds.

Next, let us compute the summation $\sum_{k=1}^{K} w_k \delta_h^k$. Denoting $n_h^k = N_h^k(x_h^k, a_h^k)$, we have:

$$\sum_{k=1}^{K} w_k \alpha_{n_h^k}^0 H = \sum_{k=1}^{K} H w_k \cdot \mathbb{I}[n_h^k = 0] \leq HSA\|w\|_\infty \;; \text{ and} \quad (C.19)$$

$$\sum_{k=1}^{K} w_k \sqrt{\frac{H^3 \iota}{n_h^k}} \overset{①}{=} O(1) \cdot \sum_{x,a} \sum_{i=1}^{N_h^K(x,a)} w_{k_i(x,a)} \sqrt{\frac{H^3 \iota}{i}}$$

$$\overset{②}{\leq} O\big(SA\|w\|_\infty + \sqrt{SA\|w\|_1 \|w\|_\infty}\big) \cdot \sqrt{H^3 \iota} \;. \quad (C.20)$$

Above,

- Equality ① is by reordering the indices $k \in [K]$ so that the ones with the same $(x, a) = (x_h^k, a_h^k)$ are grouped together; and we denote by $k_i(x, a) = k$ where $k$ is the $i$th episode where $(x, a)$ is taken at step $h$.

- Inequality ② is because $\sum_{x,a} \sum_{i=1}^{N_h^K(x,a)} w_{k_i(x,a)} = \|w\|_1$. Therefore, the left-hand side of ② is maximized when the weights are distributed to those indices $i$ that have smaller values:

$$\sum_{x,a} \sum_{i=1}^{N_h^K(x,a)} w_{k_i(x,a)} \sqrt{\frac{1}{i}} \leq \|w\|_1 + \sum_{x,a} \sum_{i=1}^{\lfloor \frac{\|w\|_1}{SA\|w\|_\infty} \rfloor} \|w\|_\infty \sqrt{\frac{1}{i}} \leq O\big(SA\|w\|_\infty + \sqrt{SA\|w\|_1 \|w\|_\infty}\big) \;.$$

To bound the second term in (C.18), which is

$$\sum_{k=1}^{K} w_k \sum_{i=1}^{n_h^k} \alpha_{n_h^k}^i \phi_{h+1}^{k_i(x_h^k, a_h^k)} \;, \quad (C.21)$$

we regroup the summands in (C.21) in a different way. For every $k' \in [K]$, we group all terms $\phi_{h+1}^{k'}$ that appear in the inner summand of (C.21)—denoting their total weight by $w'_{k'}$—and write:

$$\sum_{k=1}^{K} w_k \sum_{i=1}^{n_h^k} \alpha_{n_h^k}^i \phi_{h+1}^{k_i(x_h^k, a_h^k)} = \sum_{k'=1}^{K} w'_{k'} \cdot \phi_{h+1}^{k'} \;. \quad (C.22)$$

We make two key observations

- We have $\|w'\|_1 \leq \|w\|_1$ because $\sum_{i=1}^{t} \alpha_t^i \leq 1$.

- For every $k' \in [K]$, we note that the term $\phi_{h+1}^{k'}$ only appears on the left-hand side of (C.22) in episode $k \geq k'$, where $(x_h^k, s_h^k) = (x_h^{k'}, s_h^{k'})$. Suppose it appears in episodes $k_1', k_2', \ldots$. Then, letting $\tau = n_h^{k'}$, we have corresponding weight is $w_{k'}\alpha_\tau^\tau, w_{k_1'}\alpha_{\tau+1}^\tau, w_{k_2'}\alpha_{\tau+2}^\tau \cdots$. Therefore, the total weight satisfies

$$w'_{k'} \leq \|w\|_\infty \sum_{t=n_h^{k'}+1}^{\infty} \alpha_t^{n_h^{k'}} \leq \left(1 + \frac{1}{H}\right)\|w\|_\infty ,$$

where the final inequality uses $\sum_{t=i}^{\infty} \alpha_t^i = 1 + \frac{1}{H}$ from Lemma 4.1.c.

Plugging (C.19), (C.20), and (C.22) back into (C.18), we have:

$$\sum_{k=1}^{K} w_k \phi_h^k \leq HSA\|w\|_\infty + \sum_{k'=1}^{K} w'_{k'} \cdot \phi_{h+1}^{k'} + O\big(SA\|w\|_\infty + \sqrt{SA\|w\|_1\|w\|_\infty}\big) \cdot \sqrt{H^3\iota} ,$$

with $\|w'\|_\infty \leq (1 + \frac{1}{H})\|w\|_\infty$ and $\|w'\|_1 \leq \|w\|_\infty$. Recursing this for $h, h+1, \ldots, H$, we conclude that

$$\sum_{k=1}^{K} w_k \phi_h^k \leq O\big(SA\|w\|_\infty\sqrt{H^5\iota} + \sqrt{SA\|w\|_1\|w\|_\infty H^5\iota}\big) . \qquad \square$$

## D  Proof of Lower Bound

Recall that Jaksch et al. [10] showed that for any algorithm, there is an MDP with diameter $D$, $S$ states and $A$ actions, such that the algorithm's regret must be at least $\Omega(\sqrt{DSAT})$. The natural analogous notion of the diameter in the episodic setting is $H$, and thus this suggests a lower bound in $\Omega(\sqrt{HSAT})$, as presented in [5, 20].

We show that, in our episodic setting of this paper, one can obtain a stronger lower bound:

**Theorem 3.** *For any algorithm there exists an $H$-episodic MDP with $S$ states and $A$ actions such that for any $T$, the algorithm's regret is $\Omega(H\sqrt{SAT})$.*

This result seemingly contradicts the $O(\sqrt{HSAT})$ regret bound of Azar et al. [5]. There is no contradiction, however, because Azar et al. [5] assumes that the transition matrix $\mathbb{P}_h$ is the *same* at each step $h \in [H]$. On the contrary, in this paper we consider the more general setting where the transition matrices $\mathbb{P}_1, \ldots, \mathbb{P}_H$ are distinct for each step. Our setting can be viewed as a special case of the non-episodic MDP studied by Jaksch et al. [10], obtained by augmenting the state space to $\mathcal{S}' = \mathcal{S} \times [H]$.

Rather than providing a formal proof of Theorem 3 we give the intuition behind the construction and its analysis. The formalization itself is an easy exercise following well-known lower-bound techniques from the multi-armed bandit literature; see, e.g., [6]. For the sake of simplicity, we consider $A = 2$ and $S = 2$ (again the generalization to arbitrary $A$ and $S$ is routine).

We start by recalling the construction from Jaksch et al. [10], which we will refer to as the "JAO MDP." The reward does not depend on actions: state 1 always has reward 1 and state 0 always has reward 0. From state 1, any action takes the agent to state 0 with probability $\delta$, and to state 1 with probability $1 - \delta$. In state 0, there is one action $a^\star$ takes the agent to state 1 with probability $\delta + \varepsilon$, and the other action $a$ takes the agent to 1 with probability $\delta$. A standard Markov chain exercise shows that the stationary distribution of the optimal policy (that is, the one that in state

0 takes action $a^\star$) has a probability of being in state 1 of

$$\frac{\frac{1}{\delta}}{\frac{1}{\delta}+\frac{1}{\delta+\varepsilon}} = \frac{\delta+\varepsilon}{2\delta+\varepsilon} \geq \frac{1}{2} + \frac{\varepsilon}{6\delta} \text{ for } \varepsilon \leq \delta.$$

In contrast, acting sub-optimally (that is, taking action $a$ in state 0) leads to a uniform distribution over the two states, or equivalently a regret per time step of order $\varepsilon/\delta$. Moreover, in order to identify the two actions $a, a^\star$ (each with probability $\delta$ and $\delta + \varepsilon$), the number of observations in state 0 needs to be at least $\Omega(\delta/\varepsilon^2)$. Thus, taking the latter quantity to be $T$, one obtains the following lower bound on total regret:

$$T \times \Omega(\varepsilon/\delta) = \Omega(\sqrt{T/\delta}).$$

In the JAO MDP, the diameter is $D = \Theta(1/\delta)$. This proves the $\sqrt{DT}$ lower bound from Jaksch et al. [10].

The natural analogue of the JAO MDP for the episodic setting is to put the JAO MDP in "series" for $H$ steps (in other words, one takes $H$ steps in the JAO MDP and then restarts, say starting in state 0). The main difference with the non-episodic version is that, in $H$ steps, one may not have time to *mix*, i.e., to reach the stationary distribution over the two states. Using standard theory of Markov chains, one can show that the optimal policy on this episodic MDP has a mixing time of $\Theta(1/\delta)$. By choosing $H$ to be slightly larger than $\Theta(1/\delta)$, we have a sufficient number of steps (in each episode) to mix, and thus the previous non-episodic argument remains valid for the episodic case. This leads to a lower bound $\Omega(\sqrt{HT})$ for the episodic case, as illustrated by [5, 20].

Finally, recall that in our episodic setting, the transition matrices $\mathbb{P}_1, \ldots, \mathbb{P}_H$ may not necessarily be the same. Therefore, we can further strengthen this lower bound to $\Omega(H\sqrt{T})$ in the following way.

Let us use $H$ *distinct* JAO MDPs, each with a different optimal action $a_h^\star$, when putting them in series. In other words, for at least half of the steps $h \in H$, one has to identify the correct action $a_h^\star$ for that specific step. (If not, the per-iteration regret will again be $\Omega(\varepsilon/\delta)$.) However the number of observations in that specific step $h$ is only $T/H$, and thus one now needs to take $T/H = O(\delta/\varepsilon^2)$ (instead of $T = \Omega(\delta/\varepsilon^2)$ previously). This gives the claimed $\Omega(H\sqrt{T})$ lower bound.